\documentclass[11pt]{article}
\usepackage{amssymb, xspace, xypic,epsfig,amsmath,color}
\usepackage{subfigure}
\usepackage[all]{xy}

\newcommand\scite[1]{\cite{#1}}
\newcommand\comment[1]{}
\newcommand\eq[2]{\begin{equation}\label{#1} #2 \end{equation}}
\newcommand\eqs[2]{\begin{eqnarray}\label{#1} #2 \end{eqnarray}}

\newtheorem{thm}{Theorem}[section]

\newtheorem{df}[thm]{Definition}

\def\C{{\mathbb C}}
\def\R{{\mathbb R}}

\def\SO{{\mathcal SO}}
\def\SA{{\mathcal SA}}
\def\SE{{\mathcal SE}}
\def\O{{\mathcal O}}
\def\A{{\mathcal A}}
\def\E{{\mathcal E}}
\def\SL{{\mathcal SL}}
\def\GL{{\mathcal GL}}
\def\CK{{\mathcal K}}
\def\ijet{{\mathcal S}}

\def\p{{\bf p}}
\def\GLC2{GL(2,\C)}

\newcommand\concept[1]{\blue{\bf{#1}}}
\def\blue{\textcolor{blue} }

\newcommand\black[1]{\it{#1} }
\def\Eq#1{(\ref{#1})}

\newcommand\td[2]{\frac{d{#1}}{d{#2}}}

\newcommand\br[1]{\overline{#1}}
\newcommand{\RR}{\mathbb{R}}
\def\act{\alpha}
\def\sva{\mathcal K}
\def\zva{S}
\def\p{s} 
\onecolumn

\begin{document}
\title{Classification of curves in 2D and  3D via affine integral signatures}
\author{ S. Feng\thanks{ECE Department, North Carolina State University, Raleigh, NC, 27695-7914, E-mail: maxfeng@yahoo.com.}, I. A. Kogan\thanks{Department of Mathematics, North Carolina State University, Raleigh, NC, 27695-8205,   E-mail: iakogan@ncsu.edu. Supported  in  part  by NSF CCF-0728801 grant.}, H. Krim \thanks{ECE Department, North Carolina State University, Raleigh, NC, 27695-7914, E-mail: ahk@ncsu.edu.  Supported  in  part  by AFOSR F49620-98-1-0190 grant.}}
\maketitle

\begin{abstract}
We propose a robust classification algorithm for curves in 2D and
3D, under special and full groups of affine transformations.  To
each plane or spatial curve we assign a plane signature curve.
Curves, equivalent under an affine transformation, have the same
signature. The signatures introduced in this paper are based on
integral invariants, which behave much better on noisy images than
classically known differential invariants. The comparison with other
types of invariants is given in the introduction. Though the
integral
 invariants for planar curves were known before, the affine integral invariants for spatial curves
 are proposed here for the first time.  Using the inductive variation of the moving frame method
 we compute affine invariants in terms of Euclidean invariants. We present two types of signatures,
 the global signature and the local signature. Both signatures are independent of parameterization (curve sampling).
The global signature  depends on the choice of the initial point and
does not allow us to compare fragments of curves, and is therefore
sensitive to occlusions. The local signature, although is slightly more
sensitive to noise, is independent of the choice of the initial
point and is not sensitive to occlusions in an image. It helps
establish local equivalence of curves. The robustness of these
invariants and signatures in their application to the problem of
classification of noisy spatial curves extracted from a 3D object is
analyzed.

\end{abstract}

\section{INTRODUCTION}
\label{sect:intro}

Curves and surfaces are fundamental entities in computer vision
and pattern recognition. For example, the features of 3D or 2D
objects are often spatial or planer curves, and their
classification often reduces to  a classification of curves under
Euclidean, affine or projective transformations. A direct
comparison of curves, such as shape matching, generally requires
registration, and the ensuing complexity and difficulty in its
application  in many important problems, have recently led to a
renewed research interest in transformation invariants.

Although geometric invariants have been applied to problems in
computer image  recognition and processing for decades \cite{mundy92,mundy94,Faug94,weiss93,OST97}, designing
robust algorithms that are tolerant to noise and image occlusion
remains an open problem. We start by providing a brief overview of
various types of invariants that have appeared in computer vision
literature.   Euclidian differential invariants,  such as Euclidean
curvature and torsion for space curves, are the most classical. The
affine and projective counterparts of curvature and torsion are well
known. The dependence of curvature and torsion on high order
derivatives (up to order 3 for the Euclidean group, 6 for the affine
group and 9 for the projective group), makes numerical approximation
of these invariants highly sensitive to noise, and therefore
impractical in computer vision applications. This has motivated a
high interest in other types of invariants such as
semi-differential, or joint invariants \cite{vang92-1,
olver01,boutin00} and various types of integral invariants
\cite{sato97,hann02, lin05,man04,man06}. Integral invariants of a curve in
the latter references depend on quantities obtained by integration
of various functions along it. Since integration reduces the effect
of noise, these invariants hold a clear advantage in practical
applications.

While explicit expressions for integral invariants are known for
plane curves in 2D, they have thus far remained elusive for spatial
curves in 3D, primarily on account of their computational
complexity. With an increasing availability of 3D data acquisition
systems and subsequent emerging applications, interest in 3D
analysis and hence robust integral invariants for curves in 3D is
becoming essential.

 In \cite{feng07icassp} a hybrid integro-differential affine invariant which only uses first order
 derivatives along with integrals were computed.
Although a performance improvement over classical differential
invariants is obtained, the presence of first order derivatives
still affects the performance.

In \cite{feng07spie}, we obtain for the first time explicit formulae
of integral Euclidean and affine invariants for spatial curves in
3D. Hann and Hickman \cite{hann02} introduced and computed these for
plane curves. The type of integral invariants, computed in this
paper, may be compared with moment invariants \cite{taub92,xu06}. We
emphasize, however,  the following difference: a moment invariant
corresponds a {\it number} to a shape, whereas  an integral
invariant corresponds a {\it curve} to a curve. The standard action
of the affine group on $\R^3$ induces an action on curves. Following
the approach of \cite{hann02} we prolong this action to certain
integral expressions, called {potentials}, and then compute
invariants that depend on these integral variables. A direct
extension of \cite{hann02} to 3D, using a Fels-Olver moving frame
construction \cite{FO99} is conceptually straightforward, but the
computational complexity makes the problem intractable. An inductive
implementation of the  moving frame construction, proposed in
 \cite{kogan03}, dramatically simplifies the algebraic derivations,
as it allows one to construct invariants for the entire group from
 invariants of its subgroups: in our case affine invariants in terms
of Euclidean ones.

The integral invariants defined in \cite{hann02} and
\cite{feng07spie} are sensitive to parameterization, or sampling of
the curve in the discrete case. A uniform parameterization is
required for two curves to be compared. In order to overcome this
limitation, we propose in this paper local and global 2D/3D
signatures for the special affine and full affine group. Signatures
based on integral invariants are defined in an analogous way as
signatures based on differential  and joint invariants (see
\cite{COSTH98} for example). The global signature of a curve depends
on the choice of its initial point and does not allow a comparison
of its fragments. It is therefore sensitive to occlusions. The local
signature, although slightly more computationally involved, is
independent of the  choice of the initial point and is not sensitive
to the occlusion effects in the image. It allows to establish a
local equivalence of curves being compared.

In Section 2, after reviewing the basic facts about group actions
and invariants, we define the notion of integral jet bundle and
integral invariants. Explicit formulae for affine integral
invariants in terms of Euclidean for curves in  2D and 3D are given
in Section 3, along with their geometric interpretation.  In Section
4 we define a global integral signature which classifies curves with
a given initial point up to affine transformations. We also define a
local signature that is independent of the initial point of a curve.
In Section 5 a discrete approximation of the signature construction
is tested on curves extracted from 3D objects. The curves  are given
as discrete sequences of points, with possibly the additive noise.
The experiments show that signature construction gives a robust
method for classification of curves under affine transformations.
The method can be easily adopted to a smaller Euclidean group.

\section{Group Action and Invariants}\label{act}

In this section we review the basic terminology  for  the group
actions and invariants, as well as the concept  of prolonging the
action to jet spaces and the notion of differential invariants. We
then introduce the notion of  \emph {integral jet space} and define
the corresponding prolongation of the action which  gives rise to
\emph{ integral invariants}.
\subsection{Definitions}
\begin{df}
An action of a group $G$ on a set
  $S$ is a  map $\act\colon G\times S \to S$ that  satisfies the following two
  properties:
\begin{enumerate}
\item $\act(e,s)=s$,  $\forall s\in S$, where $e$ is the identity of the group.
 \item $\act(g_1, \act(g_2, \p))=\act(g_1\, g_2 , \p)\), for all $\p \in S$ and
$g_1,g_2\in  G$.
\end{enumerate}
\end{df}

For  $g\in G$  and  $s\in S$ we write
\(\act(g,s)=g\cdot s=\br s.\)

\begin{df}
The \emph{orbit} of a point $s\in S$ is the set $O_s=\{g\cdot s|g\in G\}$.
\end{df}
\comment{ \begin{df}
 A subset $S_1\subset S$ is  \emph{invariant } if  $g\cdot s\in
S_1$ for  $\forall s\in S_1$ and $\forall g\in G$.\footnote{ An
orbit is a smallest invariant subset. Any invariants subset is the
union of orbits.}
\end{df}
\begin{df} An action of $G$ is called \emph{ free}  if $\forall s\in S$ the isotropy group $G_s=\{g\in G|g\cdot s =s\}=\{e\}$. An action of $G$ is called \emph{ locally free}  if $\forall s\in S$ the isotropy group $G_s$ is discrete. \end{df}
}
\begin{df} A function $f\colon S\to\RR$ is called \emph{invariant} if
\eq{inf}{f(g\cdot s)=f(s),\,\forall g\in G \mbox{ and } \forall s\in
S.}
\end{df}
Invariant functions are constant along each orbit and can be used to find equivalence classes of  objects undergoing various types of transformations.

Let $\GL(n)$ denote a group of non-degenerate  $n\times n$ matrices with real entries. Its subgroup of matrices with determinant $1$ is denoted by $\SL(n)$.
The \emph{orthogonal group} is  $\O(n)=\{A\in \GL(n)|AA^T=I\}$, while the \emph{special orthogonal group}  is $\SO(n)=\{A\in \O(n)|\det A=1\}$.
The semi-direct product of $\GL(n)$   and $\RR^n$ is called the \emph{affine group}: $\A(n)=\GL(n)\ltimes \RR^n$. Its subgroup  $\SA(n)=\SL(n)\ltimes \RR^n$ is called  the \emph{special affine group}.
The \emph{Euclidean group} is $\E(n)=\O(n)\ltimes \RR^n$. Its subgroup  $\SE(n)=\SO(n)\ltimes \RR^n$ is called the \emph{special  Euclidean group}.

In the paper we consider the action of the affine group $\A(n)$ and its
subgroups on curves $\gamma(t)=(x_1(t),\dots ,x_n(t)),\, t\in[0,1]$ in $\RR^n$  by a composition of a linear transformation and a translation, for $n=2$ and $n=3$:
\eq{act2d}{
\left(
\begin{array}{c}
 \br {x_i(t)}   \\
 \vdots   \\
\br {x_n(t)}
\end{array}
\right)=
A
\left(\begin{array}{c}
  x_1 (t)  \\
\vdots  \\
 x_n(t)
\end{array}\right)
+\left(
\begin{array}{c}
  v_1  \\
 \vdots \\
  v_n
\end{array}
\right).
}
where matrix $A\in \GL(n)$ defines a linear transformations and vector $(v_1,\dots v_n)\in \R^n$ defines a translation.

\subsection{Prolongation of a group action}
 Our goal is to obtain
invariants  that classify curves up to  affine transformations. The
classical method of obtaining such invariants is  to  prolong the
action to the set of derivatives
$\{x_1^{(k)},\dots,x_i^{(k)}|k=1..l\}$ of a sufficiently high order
$l$ \eq{pract}{\br{x_i^{(1)}}(t)=\td{\br {x_i}(t)}{t},
\quad \br{x_i^{(k+1)}}(t)=\td{\br {x_i^{(k)}}(t)}{t}.}

\begin{df}Functions of $\{x_1,\dots ,x_n, x_i^{(k)}\,|\,i=1..n,\,k=1..l\}$  that are invariant under the prolonged action \Eq{pract} are called  \emph{ differential invariants} of order $l$.
\end{df}
 For  the Euclidean action on
 curves in 3D, the two lowest order invariants are called curvature and torsion, and are classically known in differential geometry. Analogous invariants for the affine  and  projective groups are also known.

 As noted in the introduction, differential invariants
 are highly sensitive to noise. We extend  the approach of \scite{hann02}  from planar curves to curves in a space of arbitrary dimension.
 Let $\gamma(t)$ parametrized  by $t\in [0,1]$ be a curve.  We define integral variables
  \eq{int}{ x_i^{[\alpha_1,\dots,\alpha_n]}(t)=\int_{0}^{t}x_1(t)^{\alpha_1}\cdots x_n(t)^{\alpha_n}dx_i(t),}
 where  the integrals are taken along the curve $\gamma(t)$ and  $\alpha_1,\dots,\alpha_n$ are non-negative integers, such that $\alpha_1+\dots+\alpha_{i-1}+\alpha_{i+1}+\dots+\alpha_{n}\neq 0$.
We call  $l=\alpha_1+\dots+\alpha_{n}$ the order of integral variables, and there are totally $n\left(\frac {(n+l)!}{n!\,l!}-(l+1)\right)$ of variables of order  less or equal to $l$.
 Integration-by-parts formula dictates certain relations among the integral variables, the  coordinates   $x_1(t),\dots x_n(t)$  of an arbitrary point   on a  curve $\gamma(t),\,t\in[0,1]$, and the  coordinates $x_1^0,\dots x_n^0$ of  the initial point  $\gamma(0)$. 
 For example  $$x_1^{[0,1,0,\dots,0]}(t)=\int_{0}^{t}x_2(t) d x_1(t)=x_2(t) x_1(t)-x_2^0 x_1^0-\int_{0}^{t}x_1(t) d x_2(t)=x_2(t)d x_1(t)-x_2^0 x_1^0-x_2^{[1,0,0,\dots,0]}(t).$$
It is not difficult to show that there are  $$N_l=(n-1)\frac {(n+l)!}{n!\,l!}-\sum_{m=1}^{n-1}\frac{(n-m+l)!}{(n-m)!\,l!}$$ {\it  independent integral  variables} of variables of order  less or equal to $l$. A canonical choice of such variables is given by:
 \eq{canonical}{ x_i^{[\alpha_{j_1},\dots,\alpha_{j_k}]}(t)=\int_{0}^{t}x_{j_1}(t)^{\alpha_{j_1}}\cdots x_{j_k}(t)^{\alpha_{j_k}}dx_i(t),\mbox { where }
j_1<j_2,\dots<j_k,\, \alpha_{j_1}>0 \mbox{ and } i>j_1.}
For example variable $x_2^{[1,0,0,\dots,0]}$ is canonical, but $x_1^{[0,1,0,\dots,0]}$ is not canonical.

 \begin{df}  Let ${\cal I}^l$  be an $N_l$-dimensional space of {independent} integral variables of order $l$ and less, then  the   \emph {integral jet space of order $l$}  (denoted $\ijet^l$) is defined to be a direct product of ${\cal I}^l$ and two copies of $\R^n$, i.e $\ijet^l={\cal I}^l\times\R^n\times\R^n$. The coordinates   $x_1,\dots x_n$ of the first copy of $\R^n$ represent an arbitrary point     on a  curve $\gamma(t),\,t\in[0,1]$,    and  coordinates $x_1^0,\dots x_n^0$  of the second copy of $\R^n$ represent  the initial point  $\gamma(0)$.\end{df}

 The action \Eq{act2d}  can be prolonged to the curves on jet space as follows:

 \eqs{Saction}{
 \nonumber \left(
\begin{array}{c}
 \br {x_i}(t)   \\
 \vdots   \\
\br {x_n}(t)
\end{array}
\right)&=&
A
\left(\begin{array}{c}
  x_1(t)   \\
\vdots  \\
 x_n(t)
\end{array}\right)
+\left(
\begin{array}{c}
  v_1  \\
 \vdots \\
  v_n
\end{array}
\right) \\
\left(
\begin{array}{c}
 \br {x_i^0}   \\
 \vdots   \\
\br {x_n^0}
\end{array}
\right)&=&
A
\left(\begin{array}{c}
  x_1^0  \\
\vdots  \\
 x_n^0
\end{array}\right)
+\left(
\begin{array}{c}
  v_1  \\
 \vdots \\
  v_n
\end{array}
\right),
\\
\nonumber \br{x_i^{[\alpha_1,\dots,\alpha_n]}}(t)&=&\int_0^t\br{x_1}^{\alpha_1}(t)\cdots \br{ x_n}^{\alpha_n}(t)d\br{x_i}(t).}
 It is important that  the integration-by-parts relations among the integral variables are respected by the prolonged action, and therefore the action on the integral jet space is $\ijet^l$ is well defined.

 \begin{df} A function on $\ijet^l$ which is \emph{ invariant} under the prolonged  action \Eq{Saction} is called \emph{integral invariant of order $l$}.
\end{df}

By introducing new variables
\eq{T}{X_i(t)=x_i(t)-x_i^0,\,i=1,\dots,n} and making the corresponding substitution into
the integrals,
we reduce the problem of finding invariants under the action
\Eq{Saction} to an equivalent but simpler problem of finding
invariant functions  of variables
  $\{X_1,\dots, X_n, X_i^{[\alpha_1,\dots,\alpha_n]}\,|\,i=1\dots n\}$ under the action of $GL(n)$ defined by
   \eqs{Sact}{
\left(
\begin{array}{c}
 \br {X_i}(t)   \\
 \vdots   \\
\br {X_n}(t)
\end{array}
\right)&=&
A
\left(\begin{array}{c}
  X_1(t)   \\
\vdots  \\
 X_n(t)
\end{array}\right)
\\
\nonumber \br{X_i^{[\alpha_1,\dots,\alpha_n]}}(t)&=&\int_0^t \br{X_1}^{\alpha_1}(t)\cdots \br{ X_n}^{\alpha_n}(t)\,d\br{X_i}(t).}
  Invariants with respect to \Eq{Saction} may be obtained from invariants with respect to \Eq{Sact} by making substitution \Eq{T}.\footnote{
    This reduction by the group of  translations can be put in the context of inductive method described in  Appendix. We feel, however, that making this step ``upfront" makes the presentation more transparent.}
Invariants with respect to a very general class of  actions of continuous finite-dimensional groups    on  manifolds  can be computed using 
Fels-Olver generalization
\cite{FO99} of Cartan's moving frame method (see also
 its algebraic reformulation  \cite{hk:focm}). The method consists of choosing a cross-section to the orbits and finding the coordinates of the projection  along the orbits of a generic point on a manifold  to the cross-setion (see Appendix for more details).  It can be, in theory, applied to find the invariants under the  action  described by \Eq{Sact} for arbitrary $n$. Hann and Hickman  \cite{hann02}  used Fels-Olver method to compute integral invariants   for \emph{planar
curves} ($n=2$) under affine transformations and a certain subgroup of
projective transformations. The corresponding derivation of
invariants for \emph{spatial curves}   ($n=3$) remained, however, out of reach
due to computational complexity (it is often the case in the computational invariant theory that practical computations become unfeasible as the dimension of the group increases, despite the availability of a theoretical method to compute them \cite{St93, DK02}.) In \cite{feng07spie}, we
derived, for the first time, integral invariants  under the Euclidean and affine transformations for \emph{spatial curves}  using an inductive variation of the moving
frame method \scite{kogan03}, which 
allowes one to construct invariants for the entire group in terms of
invariants of its subgroups: in our case, affine invariants in terms
of Euclidean.  Explicit derivation of invariants  for curves  of higher in the space of higher dimension ($n>3$) remains an open problem, which seems at present, to be of more theoretical, than of practical interest.  


\section{Integral invariants in 2D and 3D}
In this section we present explicit formulas for integral invariants  for $n=2$ (plane curves) and $n=3$ (spatial curves) under the affine  action \Eq{Saction}.  The affine invariants are written in terms of the Euclidean invariants. We 
 discuss their properties and geometric interpretation.  The
inductive derivation of these invariants is outlined in  the Appendix.

\subsection{ Integral Affine Invariants for Curves in 2D}
\label{iinv2d} The  standard affine group action on
curves in $\mathbb{R}^{2}$:
\begin{equation*}\left(
 \begin{array}{c}
  \br{x}(t)\\
  \br{y}(t)\\
\end{array}%
\right)=\left(%
\begin{array}{ccc}
  a_{11} & a_{12}  \\
  a_{21} & a_{22} \\
 \end{array}%
\right)\left(\begin{array}{c}
  x(t)\\
  {y}(t)\\
\end{array}\right)+\left(\begin{array}{c}
  v_1\\
 v_2\\
\end{array}\right),\quad \det\left(%
\begin{array}{cc}
  a_{11} & a_{12} \\
  a_{21} & a_{22} \\
\end{array} %
\right)\neq 0,
\end{equation*}
 prolongs to the action on integral variables
 up to the third order. 
 
 By translating
the initial point $\gamma(0)$ to the origin  and making the
corresponding  substitution $X(t)=x(t)-x(0), Y(t)=y(t)-y(0)$  in
the integrals, we reduce the problem to computing invariants under the action \Eq{Sact} with $n=2$. 
Among 12 integral variables
 \eqs{int2D}{\nonumber X^{[i,j]}(t)&=&\int_{0}^{t}X(t)^{i}Y(t)^{j}dX(t),\quad j\neq 0, i+j\leq 3\\ Y^{[i,j]}(t)&=&\int_{0}^{t}X(t)^{i}Y(t)^{j}dY(t), \quad i\neq 0, i+j\leq 3}
   we  make a canonical choice of  6 independent:  $Y^{[1,0]}, Y^{[2,0]},
 Y^{[1,1]}, Y^{[3,0]}, Y^{[2,1]}, Y^{[1,2]}, $ as suggested by formula   \Eq{canonical}. The rest can be expressed in terms of those using integration by parts formulas, as follows:
  \eqs{2D-by-parts}{
  \nonumber X^{[0,1]}&=&XY-Y^{[1,0]},\\
  \nonumber  X^{[0,2]}&=& X\,Y^2-2\,Y^{[1,1]},\\
    \nonumber  X^{[1,1]}&=& \frac 1 2 X^2\,Y-\frac 1 2\,Y^{[2,0]},\\
  X^{[1,2]}&=&\frac 12 X^2\,Y^2-Y^{[2,1]},\\
   \nonumber  X^{[0,3]}&=& Y^3X-3Y^{[1,2]},\\
      \nonumber X^{[2,1]}&=& \frac 1 3 X^3Y-\frac 1 3Y^{[3,0]}.
  }
  This reduces the problem to   finding  invariants under the following  $GL(2)$-action on $\R^8$.
   Denote $\det A:=a_{11}a_{22}-a_{21}a_{21}$. The action is defined by the following equations:
  \eqs{2D}{ \nonumber
 \br {X}&=& a_{11}X+a_{12}Y,\qquad  \br {Y}   = a_{21}X + a_{22}Y,
 \\
 \nonumber \br{{Y^{[1,0]}}}&=&(\det A)\, {Y^{[1,0]}}+\frac 1 2 a_{11}a_{21}{X^2}
 +a_{12}a_{21}XY+\frac 12 a_{12}a_{22} {Y^2},\\
\nonumber  \br{Y^{[1,1]}}
 &=&(\det A) \left(a_{{22}}Y^{[1,1]} +\frac 12  a_{{21}}Y^{[2,0]}\right)+\frac1 3{{a_{{21}}}^{2}a_{{11}}{X}^{3}}
+\frac 12 a_{21} (a_{11}a_{22}+a_{12}a_{21}){X}^{2}Y\\
\nonumber &&+ 
 a_{{21}}a
_{{12}}a_{{22}}X{Y}^{2}+\frac 1 3{a_{{22}}}^{2}a_{{12}}{Y}^{3}\\
\nonumber  \br{Y^{[2,0]}} &=&(\det A)\left(a_{{11}}Y^{[2,0]}+2\,a_{{12}}Y^{[1,1]} \right)+ \frac 1 3{a_{{11}}}^{2}a_{{21}}{X}^{3}+a_{{11}}a_{{12}}a_{{21}}{X}^{
2}Y+a_{12}^2a_{21}X{Y}^{2}\\
\nonumber && +\frac 1 3\,a_{11}a_{12}a_{21}{Y}^{3},\\
\br  {Y^{[1,2]}}&=&(\det A)\left(a_{22}^{2}\,Y^{[1,2]}+\frac 1 3 \,a_{21}^{2}\,{
Y^{[3,0]}}+a_{21}a_{22}\,{
Y^{[2,1]}}\right)\\
\nonumber &&+\frac{1}{4}a_{{11}}{a_{{21}}}^{3}{X}^{4}+ \frac 1 3 a_{21}^2\,(2a_{11}a_
{22}+a_{12}a_{21})\,X^3 Y\\
 \nonumber & &  + \frac 1 2 a_{21}a_{22}\,(2\,a_{12}a_{21
}+a_{11}a_{22})\, 
X^2Y^2 +a_{{12}}a_{{21}}{a_{{22}}}^{2}\,
XY^3\,+\frac{1}{4}a_{{12}}{a_{{22}}}^{3}{Y}^{4}
,\\
  \nonumber\br {Y^{[2,1]}}&=&   (\det A)\left((a_{11}a_{22}+ 
a_{12}a_{21})Y^{[2,1]}+2\,a_{12}a_{22}{ Y^{[1,2]}} +\frac 2 3 \,a_{11}a_{21}
Y^{[3,0]}\right)\\
\nonumber & &+\frac{1}{4}\,a_{11}^{2}a_{21}^{2}\, {X}^{4}
+\frac 13 a_{11}
\,a_{ 21}(a_{11}a_{22}+2\,a_{12}a_{21}) X^3\,Y+\frac  1 2 a_{12}a_{21}(2\,a_{11}a_{22}+ a_{12}
a_{21})X^2Y^2
\\
\nonumber & &\,+a_{12}^{2}a_{21}a_{22}\,XY^3+ \frac{1}{4}{a_{{12}}}^{
2}{a_{{22}}}^{2}{Y}^{4}
, \\
 \nonumber \br{Y^{[3,0]}}&=&(\det A)\left(a_{11}^2\,Y^{[3,0]}+3\,a_{12}^2{ Y^{[1,2]}}+3\,a_{11}a_{12}
Y^{[2,1]}\right)\\
\nonumber & &\,+\frac{1}{4}a_{11}^3a_{{21}}\,{X}^{4}+a_{11}^{2}a_{12}a_
{21}\,X^3Y+\frac 3 2\, a_{11}a_{12}^2a_{21} X^2Y^2
+a_{12}^{3}a_{21}\,Y^3X+\frac 1 4\,a_{12}^{3}a_{22}\,{Y}^{4}.
 }

We  restrict the  above action  to  the subgroup  $\SO(2)$ of
rotation matrices by setting $a_{11}=\cos\phi,a_{12}= -\sin\phi,
a_{21}=\sin\phi, a_{22}=\cos\phi$. We use the moving frame method to
find invariants as described in the Appendix.
 Computationally this reduces to the substitution  $a_{11}=\frac X r,a_{12}= \frac Y r,
a_{21}=\frac {-Y} r, a_{22}=\frac X r$, where $r= {\sqrt{X^2+Y^2}}$  in \Eq{2D}. The resulting non-constant expressions comprise a set of  generating invariants for the $SO(2)$ action:

 \eqs{SO2inv}{
\nonumber  X_{\SE}&=& {\sqrt{X^2+Y^2}}=r,\\
\nonumber  Y_{\SE}^{[1,0]}&=&Y^{[1,0]}-\frac {XY}2,\\
{Y^{[1,1]}_{\SE}}&=& \frac  1 r  \left({Y^{[1,1]}\,X-\frac 12 Y^{[2,0]}\,Y-\frac 1 6\,X^2Y^2}\right), \\ 
\nonumber Y^{[2,0]}_{\SE}&=& \frac  1 r \left(Y^{[2,0]}\,X+2\,Y^{[1,1]}\,Y-\frac  13\,X^3Y -\frac 23 XY^3\right),\\
\nonumber Y^{[1,2]}_{\SE}&=&\frac 1 {r^2}\left(
Y^{[1,2]}\,X^2-Y^{[2,1]}\,XY+\frac 13 
Y^{[3,0]}\,Y^2 -\frac 1 {12} X^3Y^3\right),  \\
\nonumber Y^{[2,1]}_{\SE}&=&\frac{1}{r^2}\left(Y^{[2,1]}\,(X^2-Y^2)+2\,Y^{[1,2]}\,XY+\frac 2 3Y^{[3,0]}\,XY-\frac 1 4 X^2Y^4-\frac 1 {12}X^4Y^2\right)
, \\
 \nonumber Y^{[3,0]}_{\SE}&=&\frac{1}{r^2}\left(Y^{[3,0]}\,X^2+3\,Y^{[1,2]}\,Y^2+3\,Y^{[2,1]}\,XY-\frac 1 4 X^5Y-\frac 3 4 X^3Y^3-\frac 3 4 XY^5\right).
 }
 The  invariants with respect to the special
Euclidean group are obtained  by making a substitution of $Y= y-y^0$
and  $X= x-x^0$ in the above expressions \Eq{SO2inv}: \footnote{The notation
for invariants suggests a certain correspondence between the
invariants and the coordinate functions of the integral jet space,
which we make clear in the Appendix.}
We note that since the denominators in the above formulas are
invariant, the numerators are also invariant.

We use the inductive approach, described in the Appendix,  to build   invariants under the
$SL(2)$-action defined by Eq.\Eq{2D} with the condition
$\det A=1.$ The inductive method yields $\SA(2)$-invariants  in
terms of $\SE(2)$-invariants \Eq{SO2inv}:

\eqs{SL2inv} {\nonumber Y^{[1,0]}_{\SA}&=&Y^{[1,0]}_{\SE}=Y^{[1,0]}-\frac {XY}2, \\
\nonumber Y^{[1,1]}_{\SA}&=&X_{\SE}Y^{[1,1]}_{\SE}={Y^{[1,1]}\,X-\frac 12 Y^{[2,0]}\,Y-\frac 1 6\,X^2Y^2},
\\
 Y^{[1,2]}_{\SA}&=&Y^{[1,2]}_{\SE}X_{\SE}^2=Y^{[1,2]}\,X^2-Y^{[2,1]}\,XY+\frac 13 
Y^{[3,0]}\,Y^2 -\frac 1 {12} X^3Y^3,
 \\
\nonumber Y^{[2,1]}_{\SA}&=&Y^{[2,1]}_{\SE}- \frac{Y^{[2,0]}_{\SE}}{Y^{[1,1]}_{\SE}}Y^{[1,2]}_{\SE}
,\\ 
 {Y^{[3,1]}_{\SA}}&=&\frac 1{X_{\SE}^2}\left(Y^{[3,0]}_{\SE}+\frac 3 2 \frac{Y^{[2,0]}_{\SE}}{Y^{[1,1]}_{\SE} }Y^{[2,1]}_{\SE}+\frac 3 4\left(\frac{Y^{[2,0]}_{\SE}}{Y^{[1,1]}_{\SE}} \right)^2Y^{[1,2]}_{\SE}\right)
  }By replacing $(X,Y)$ with  $(x-x^0,y-y^0)$ in Eq.\Eq{SL2inv} we return to the integral jet space coordinates.   In particular, $Y^{[1,0]}_{\SA}=Y^{[1,0]}-\frac  12 {XY}=\int_0^t \left(x-x^0\right)dy-\frac 1 2 (x-x^0)(y-y^0)$.

The following three special affine invariants
are  used in the next section to solve the classification  problem
with respect to  both special and full affine groups:
\eqs{2dinv}{\nonumber I_1&=&Y^{[1,0]}_{\SA}=Y^{[1,0]}-\frac  12 {XY},\\
I_2&=&Y^{[1,1]}_{\SA}={Y^{[1,1]}\,X-\frac 12 Y^{[2,0]}\,Y-\frac 1 6\,X^2Y^2},
\\
\nonumber I_3&=& Y^{[1,2]}_{\SA}=Y^{[1,2]}\,X^2-Y^{[2,1]}\,XY+\frac 13 
Y^{[3,0]}\,Y^2 -\frac 1 {12} X^3Y^3.}

To obtain invariants with respect to the full affine group we need
to consider the effect of reflections and arbitrary scaling on the
above invariants. We note that the transformation $x\to \lambda x$
and    $y \to-\lambda y$ induces the transformation
$I_1\to-\lambda^2 I_1$,  $I_2\to \lambda^4 I_2$ and $I_3\to
-\lambda^6 I_3$. The following rational expressions are thus
invariant with respect to the full affine group: \eqs{GL2inv}{
\nonumber I_2^{\A}&=&\frac{I_2}{I_1^2}=\frac{Y^{[1,1]}\,X-\frac 12 Y^{[2,0]}\,Y-\frac 1 6\,X^2Y^2}{({Y^{[1,0]}}-\frac 12 XY)^2},\\
I_3^{\A}&=&\frac{I_3}{I_1^3}=\frac{Y^{[1,2]}\,X^2-Y^{[2,1]}\,XY+\frac 13 
Y^{[3,0]}\,Y^2 -\frac 1 {12} X^3Y^3}{({Y^{[1,0]}}-\frac 12 XY)^3}}
The first of the above invariants is equivalent to the one obtained in \cite{hann02}.

\subsection{Geometric Interpretation of Invariants for Plane Curves}

\begin{figure}[h]
\centerline{ \epsfig{figure=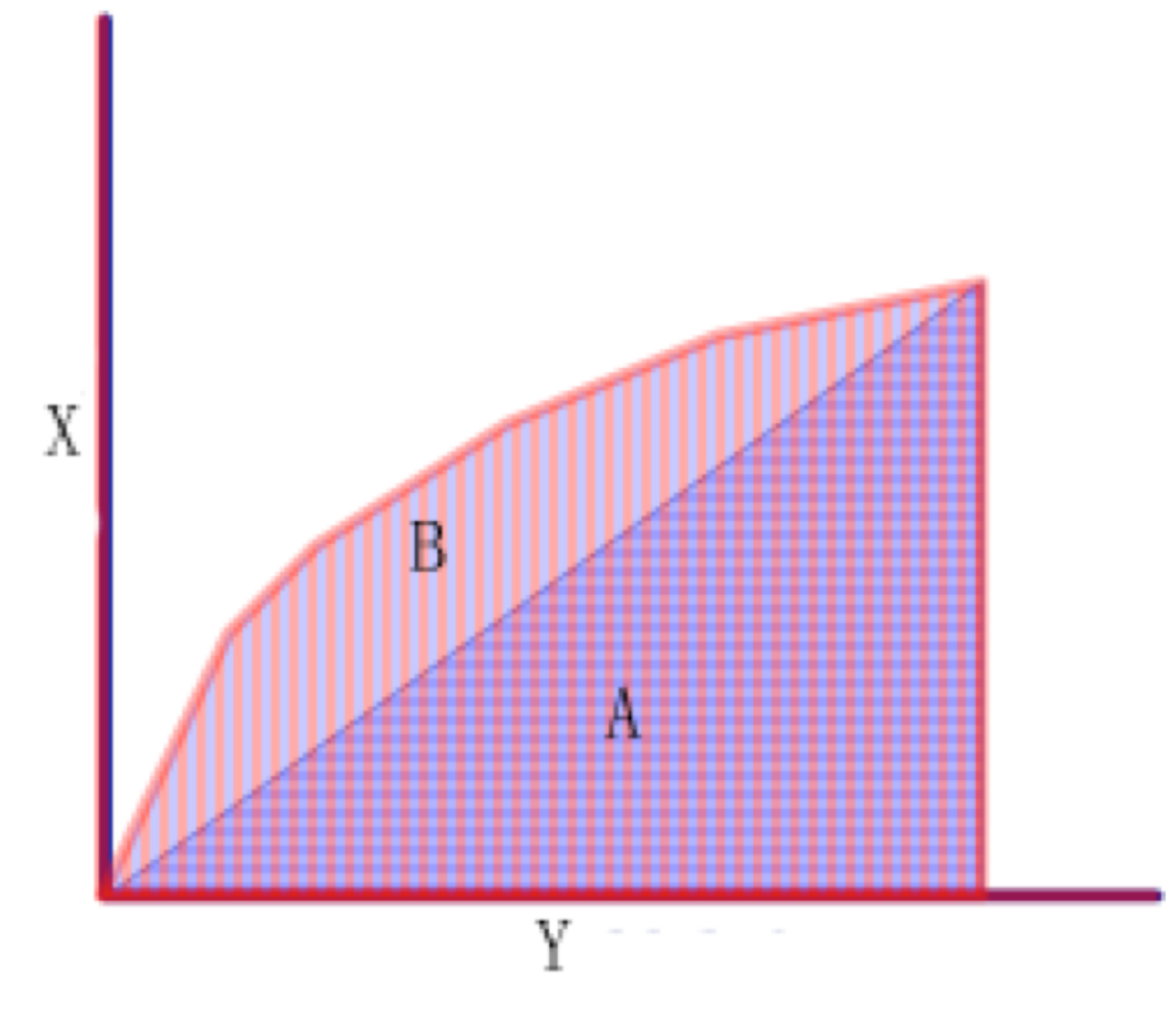,height=2in} }
\caption{{\it \small Geometric interpretation of  the invariants $I_1$}}
\label{fig:3dgeo1}
\end{figure}

The first two integral invariants \Eq{2dinv} readily lend themselves to a
geometric interpretation. Invariants $I_1$ is the signed area $B$ between the curve segment  and the secant (see Figure~\ref{fig:3dgeo1}).
Indeed, the term $Y^{[1,0]}$ in the invariant
$I_1$ is  the signed area  between the curve $\gamma(t)$ (whose initial point is translated to the origin) and the $Y$-axis,  while $\frac{XY}{2}$ is the signed area of
the triangle $A$. Their difference is the area $B$.   Since  the $\SA(2)$- action preserves areas, $I_1$  is clearly  an invariant.
\begin{figure}[h]
\centerline{ \epsfig{figure=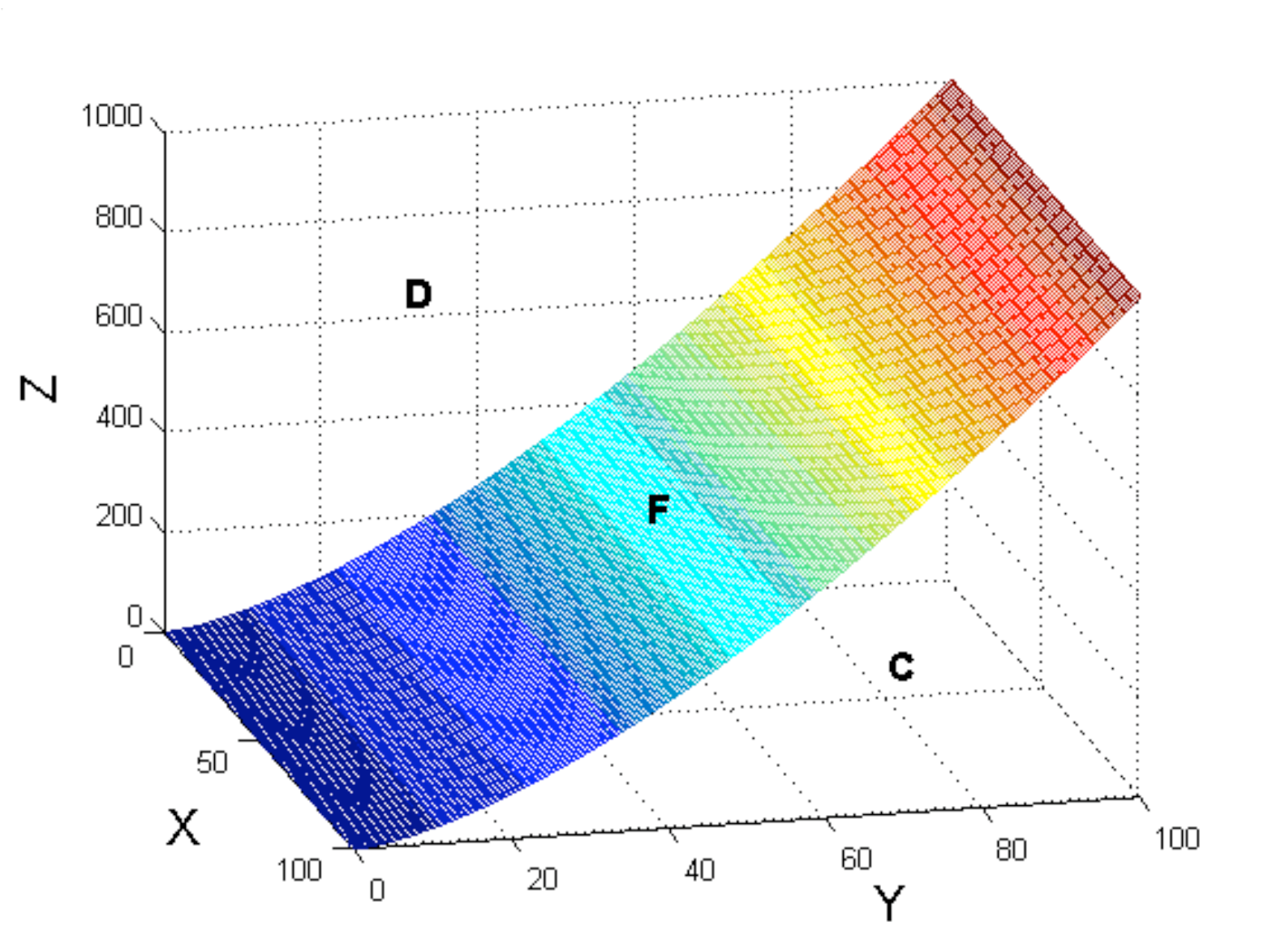,height=2in} }
\caption{{\it \small Geometric interpretation of the invariants  $I_2$}}
\label{fig:3dgeo2}
\end{figure}

The interpretation of $I_2$ is slightly more subtle. Using that $Y^{[2,0]}=X^2Y-2X^{[1,1]}$ 	and rearranging the
terms we rewrite $I_2$ as  \eq{2DI2}{ I_2=-\frac
13(({X}^{2}{Y}^{2}-3X\,Y^{[1,1]})+({X}^{2}{Y}^{2}-3Y\,X^{[1,1]})), \mbox{
where }X=x-x^0, Y=y-y^0.} Further, the  curve $\gamma(t)$ is lifted
from 2D to 3D by defining $z(t)=x(t)y(t)$ (similarly to the kernel
idea), and \Eq{2DI2} is rewritten as \eq{2DI2f}{I_2 =
-\frac{1}{3}(({X}{Y}Z-3X\int_{0}^t {Z}d{Y})+({X}{Y}Z-3Y\int_{0}^t
{Z}d{X}))\mbox{ where } Z=XY=(x-x^0)(y-y^0).}

The geometric meaning of
$({X}{Y}Z-3X\int_{0}^t {Z}d{Y})$ is illustrated in
Figure~\ref{fig:3dgeo2}. The term $\int_{0}^t {Z}d{Y}$ is  the
signed area  ``under'' the  plane curve $(Y(t), Z(t))$ in the $YZ$-plane.
Thus $X\int_{0}^t {Z}d{Y}$ is the signed volume C under the surface
$F=\gamma(t)\times [0,X(t)]$ in Figure~\ref{fig:3dgeo2}. Since $XYZ$ is the signed volume of a
rectangular prism (C+D in Figure~\ref{fig:3dgeo2}), then
${X}{Y}Z-3X\int_{0}^t {Z}d{Y}$ is the signed volume of the
rectangular prism (C+D) minus three times the volume C ``under'' the
surface $\gamma(t)\times[0,X(t)]$. Interchanging $X$
and $Y$ we obtain a similar interpretation for ${X}{Y}Z-3Y\int_{0}^t
{Z}d{X}$.

\subsection{ Integral Affine Invariants for  Curves in 3D}
The  standard affine group action on  curves in $\mathbb{R}^{3}$:
\begin{equation*}\left(
 \begin{array}{c}
  \br{x}(t)\\
  \br{y}(t)\\
  \br{z}(t)
\end{array}%
\right)=\left(%
\begin{array}{ccc}
  a_{11} & a_{12}& a_{13}  \\
  a_{21} & a_{22} & a_{23}\\
  a_{31} & a_{32} & a_{33}
 \end{array}%
\right)\left(\begin{array}{c}
  x(t)\\
  {y}(t)\\
  z(t)
\end{array}\right)+\left(\begin{array}{c}
  v_1\\
 v_2\\
  v_3
\end{array}\right),\quad \det\left(%
\begin{array}{ccc}
 a_{11} & a_{12}& a_{13}  \\
  a_{21} & a_{22} & a_{23}\\
  a_{31} & a_{32} & a_{33}
\end{array} %
\right)\neq 0,
\end{equation*}
 prolongs  to the action to integral
variables
 up to second order.
 We translate the initial point $\gamma(0)$ to the origin, and
make the corresponding  substitution $X(t)=x(t)-x(0),
Y(t)=y(t)-y(0), Z(t)=z(t)-z(0)$  in the integrals. This reduces the
problem to computing $SL(3)$-invariants under the action \Eq{Sact} with $n=3$. 
Among 21 integral variables
 \eqs{int3}{\nonumber X^{[i,j,k]}(t)&=&\int_{0}^{t}X(t)^{i}Y(t)^{j}Z(t)^{k}dX(t),\quad j+k\neq 0, i+j+k\leq 2,\\
  Y^{[i,j,k]}(t)&=&\int_{0}^{t}X(t)^{i}Y(t)^{j}Z(t)^{k}dY(t), \quad i+k\neq 0, i+j+k\leq 2,\\
 \nonumber Z^{[i,j,k]}(t)&=&\int_{0}^{t}X(t)^{i}Y(t)^{j}Z(t)^{k}dZ(t), \quad i+j\neq 0, i+j+k\leq 2,}
we  choose 11 independent:
$X^{[1,1,0]},X^{[1,0,1]},X^{[0,2,0]},Z^{[1,0,0]},Y^{[1,0,0]},Y^{[1,0,1]},Z^{[0,1,0]},Z^{[0,1,1]},Z^{[0,2,0]},Z^{[1,0,1]},Z^{[1,1,0]}.$
\footnote{The canonical choice dictated by \Eq{canonical} is $Y^{[1,0,0]},Y^{[2,0,0]},Y^{[1,1,0]},Y^{[1,0,1]}, Z^{[1,0,0]},Z^{[0,1,0]},Z^{[0,1,1]},Z^{[0,2,0]},Z^{[2,0,0]},Z^{[1,0,1]},Z^{[1,1,0]}. $ We made a computation with an equivalent but non-canonical set of variables.}
The rest can be expressed in terms of those using the integration-by-parts formula.  Using the  inductive
approach, we first compute the invariants  with respect to rotations
$SO(3)$. We find  the following $8$ independent invariants.  We
obtain  $SE(3)$-invariants by replacing $(X,Y,Z)$ with
$(x-x^0,y-y^0,z-z^0)$ ( See Appendix for details of the  derivation.)

\eqs{SO3inv}{\nonumber X_\SE&=&\sqrt{X^2+Y^2+Z^2},\\
\nonumber Z^{[0,1,0]}_\SE&=&{\frac {XYZ-2\,XZ^{[0,1,0]}+2\,YZ^{[1,0,0]}-2\,ZY^{[1,0,0]}}{2\sqrt {{X}^
{2}+{Y}^{2}+{Z}^{2}}}},\\
Y^{[1,0,0]}_\SE&=&{\frac {{ Z^{[0,2,0]}_R}\,{ Y^{[1,0,0]}_R}+2\,{ Z^{[0,1,1]}_R}\,{ Z^{[1,0,0]}_R}}{
\sqrt {{{ Z^{[0,2,0]}_R}}^{2}+4\,{{ Z^{[0,1,1]}_R}}^{2}}}},\\
   \nonumber   Y^{[1,0,1]}_\SE&=&-{\frac {-2\,{ Z^{[0,2,0]}_R}\,{ Z^{[0,1,1]}_R}\,{ Z^{[1,0,1]}_R}-{ Z^{[0,1,1]}_R}\,{ Z^{[0,2,0]}_R}
\,{ X^{[0,2,0]}_R}+4\,{{ Z^{[0,1,1]}_R}}^{2}{
Z^{[1,1,0]}_R}-{{ Z^{[0,2,0]}_R}}^{2}{ Y^{[1,0,1]}_R}} {{{
Z^{[0,2,0]}_R}}^{2}+4\,{{ Z^{[0,1,1]}_R}}^{2}}}
,\\
   \nonumber Z^{[0,2,0]}_\SE&=&\sqrt {{{ Z^{[0,2,0]}_R}}^{2}+4\,{{ Z^{[0,1,1]}_R}}^{2}}
,\\
 \nonumber Z^{[1,0,1]}_\SE&=&-{\frac {2\,{ Z^{[0,2,0]}_R}\,{ Z^{[0,1,1]}_R}\,{ Z^{[1,1,0]}_R}-{{ Z^{[0,2,0]}_R}}^{2}{
Z^{[1,0,1]}_R}+2\,{{ Z^{[0,1,1]}_R}}^{2}{ X^{[0,2,0]}_R}+2\,{
Z^{[0,1,1]}_R}\,{ Z^{[0,2,0]}_R}\,{
Y^{[1,0,1]}_R}}{{{ Z^{[0,2,0]}_R}}^{2}+4\,{{ Z^{[0,1,1]}_R}}^{2}}},\\
   \nonumber Z^{[1,1,0]}_\SE&=&{\frac {2\,{ Z^{[0,2,0]}_R}\,{ Z^{[0,1,1]}_R}\,{ Z^{[1,0,1]}_R}+{ Z^{[0,1,1]}_R}\,{ Z^{[0,2,0]}_R}\,
{ X^{[0,2,0]}_R}-4\,{{ Z^{[0,1,1]}_R}}^{2}{ Y^{[1,0,1]}_R}+{{
Z^{[0,2,0]}_R}}^{2}{ Z^{[1,1,0]}_R}}{{ { Z^{[0,2,0]}_R}}^{2}+4\,{{
Z^{[0,1,1]}_R}}^{2}}},} where expressions
$Z^{[1,0,0]}_R,Z^{[0,1,0]}_R,Y^{[1,0,0]}_R,Z^{[0,1,1]}_R,Z^{[0,2,0]}_R,Z^{[1,0,1]}_R,Z^{[1,1,0]}_R,Y^{[1,0,1]}_R,X^{[1,1,0]}_R,X^{[1,0,1]}_R,X^{[0,2,0]}_R$
are provided at the end of the Appendix.

We use them to construct the following invariants with respect to the $SA(3)$ action.
\eqs{SL3inv}{   \nonumber X_\SA&=& Z^{[0,1,0]}_{\SE}X_{\SE},\\
 Y_\SA^{[1,0,1]}&=&
\frac{2Y_\SE^{[1,0,1]}Z_\SE^{[0,1,0]}-2Z_\SE^{[0,1,0]}Z_\SE^{[1,1,0]}+3Z_\SE^{[0,2,0]}Z_\SE^{[1,0,0]}}{2Z_\SE^{[0,1,0]}}-\frac{1}{2},\\
   \nonumber Z_\SA^{[1,0,1]} &=&\frac{Z^{[1,0,1]}_\SE {Z_\SE^{[0,2,0]}}^2}{{Z_\SE^{[0,1,0]}}^3}.
} We introduce a simpler notation for the special affine invariants
which will subsequently be used to solve the classification  problem
with respect to  both the special and the full affine groups:
\eqs{3dinv}{\nonumber J_1&=&\,X_{\SA}=n_{1}X+n_{2}Z-n_{3}Y,\\
\nonumber J_2&=&-4\left(Y^{[1,0,1]}_{\SA}+\frac 1 2\right)X_{\SA}= 2\,n_{{2}} (
XY{Z}^{2}-3\,Z^{[{0,1,1}]}X+3\,YZ^{[{1,0,1}]}
-ZZ^{[{1,1,0}}-2\, ZY^{[{1,0,1}]}) \\
 &+&n_{{3}} ( 2\,X{Y}^{2}Z+3\,XZ^{[{0,2,0}]} -3\,ZX^{[{
0,2,0}]}-4\,YZ^{[{1,1,0}]}
-2\,YY^{[{1,0,1}]} )\\
\nonumber&-&2\,n_{{1}} ( 3\,YX^{[{1,0,1}]}
-3\,ZX^{[{ 1,1,0}]}+XZ^{[{1,1,0}]}-XY^{[{1,0,1}]} ),\\
\nonumber J_3&=& \frac{27}{8}Z^{[1,0,1]}_{\SA}X_{SA}^3,} where
$X=x-x^0,Y=y-y^0, Z=z-z^0$ and $n_{{1}}=\frac {YZ} 2-\,Z^{{[0,1,0]}},
n_{{2}}=\frac{XY} 2-\,Y^{{[1,0,0]}}, n_{{3}}=\frac {XZ}
2-\,Z^{{[1,0,0]}}.$ The expression of the third invariant in terms of
the original integral variables is too long to be included.

 To obtain the invariants with respect to the full affine group we need to consider the effect of  reflection and scaling on these invariants. For $\lambda\in \R$ scaling $(x,y,z)\to( \lambda x,\lambda y,-\lambda z)$ induces scaling
  $J_1\to -\lambda^3 J_1,\, J_2\to \lambda^6 J_2 $ and $J_3\to \lambda^6 J_3$.
 We therefore obtain  the following two invariants with respect to the full affine group of transformations:
\eq{GL3inv}{
 J_2^\A=\frac {J_2} {J_1^2} \mbox{ and } J_3^\A=\frac {J_3} {J_1^2}.
}

\subsection{Geometric Interpretation of Invariants for Spatial Curves}
The first invariant $J_{1}$ may be viewed as an extension of the
2D invariant $I_{1}$ to 3D. Indeed, $n_{1}$, $n_{2}$, and
$n_{3}$ represent exactly the same area as the 2D invariant
$I_{1}$(in Figure\ref{fig:3dgeo1}) in three coordinate planes. They
are extended from 2D area to 3D volume by multiplying by $X$, $Z$,
and $Y$ respectively. For example, $n_{1}X$ is the volume $C$
under surface $F$ in Figure~\ref{fig:3dgeo2}, and $n_{2}Z$ and
$n_{3}Y$ are similar volumes obtained by relabelling of $X$, $Y$,
$Z$ axis. Therefore, the invariant $J_{1}$ is  the summation of
two volumes $n_{1}X$ and $n_{2}Z$ minus the volume $n_{3}Y$.
 The
geometric interpretation of the  invariants $J_2$ and $J_3$,
however, remains at the present time unclear to us.

\section{Curve Classification via Integral Signatures}
\begin{figure}[h]\caption{} \centerline{
\begin{tabular}{ccc}
\epsfig{figure=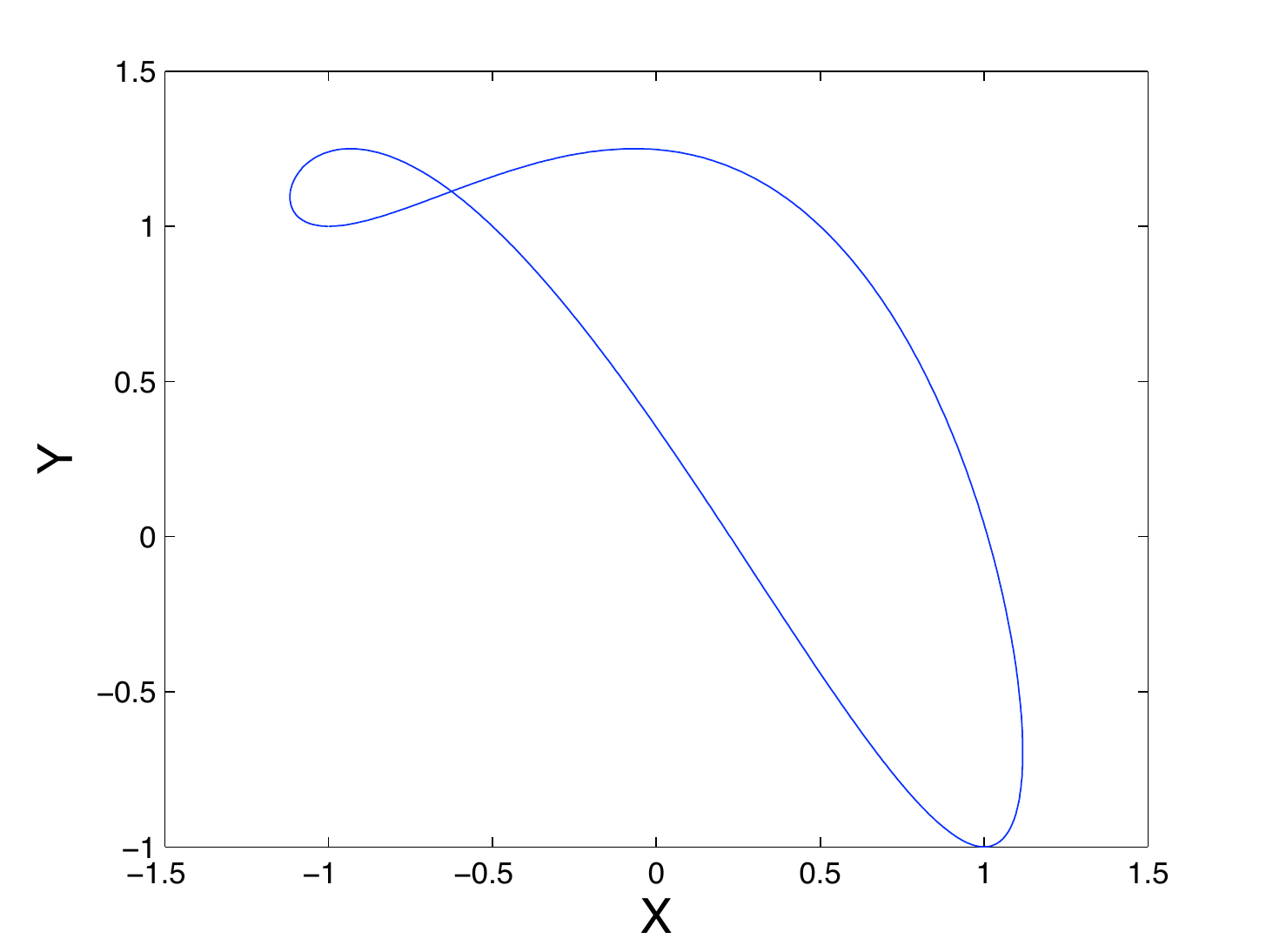,height=1.5in} &
\epsfig{figure=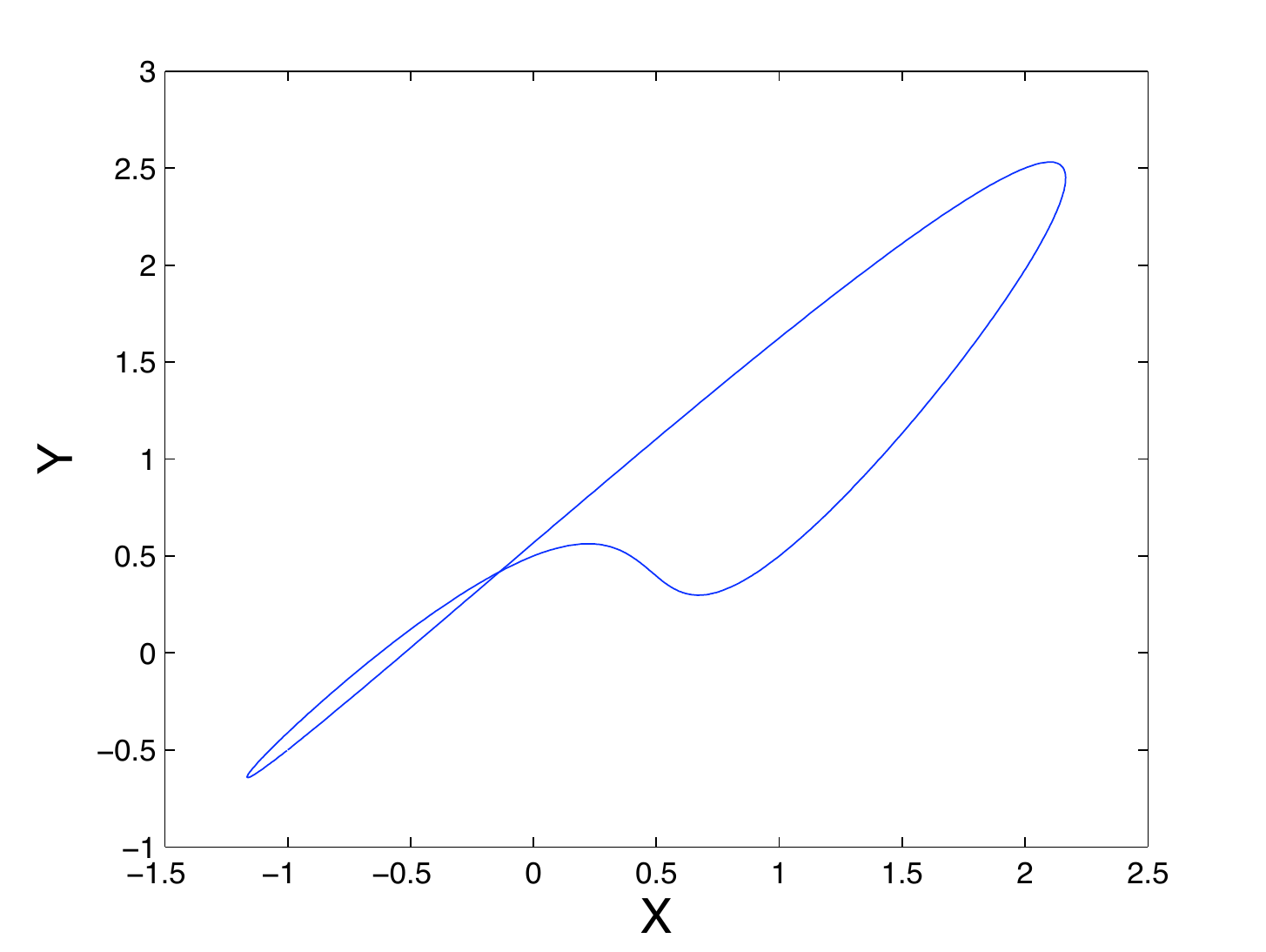,height=1.5in} &\epsfig{figure=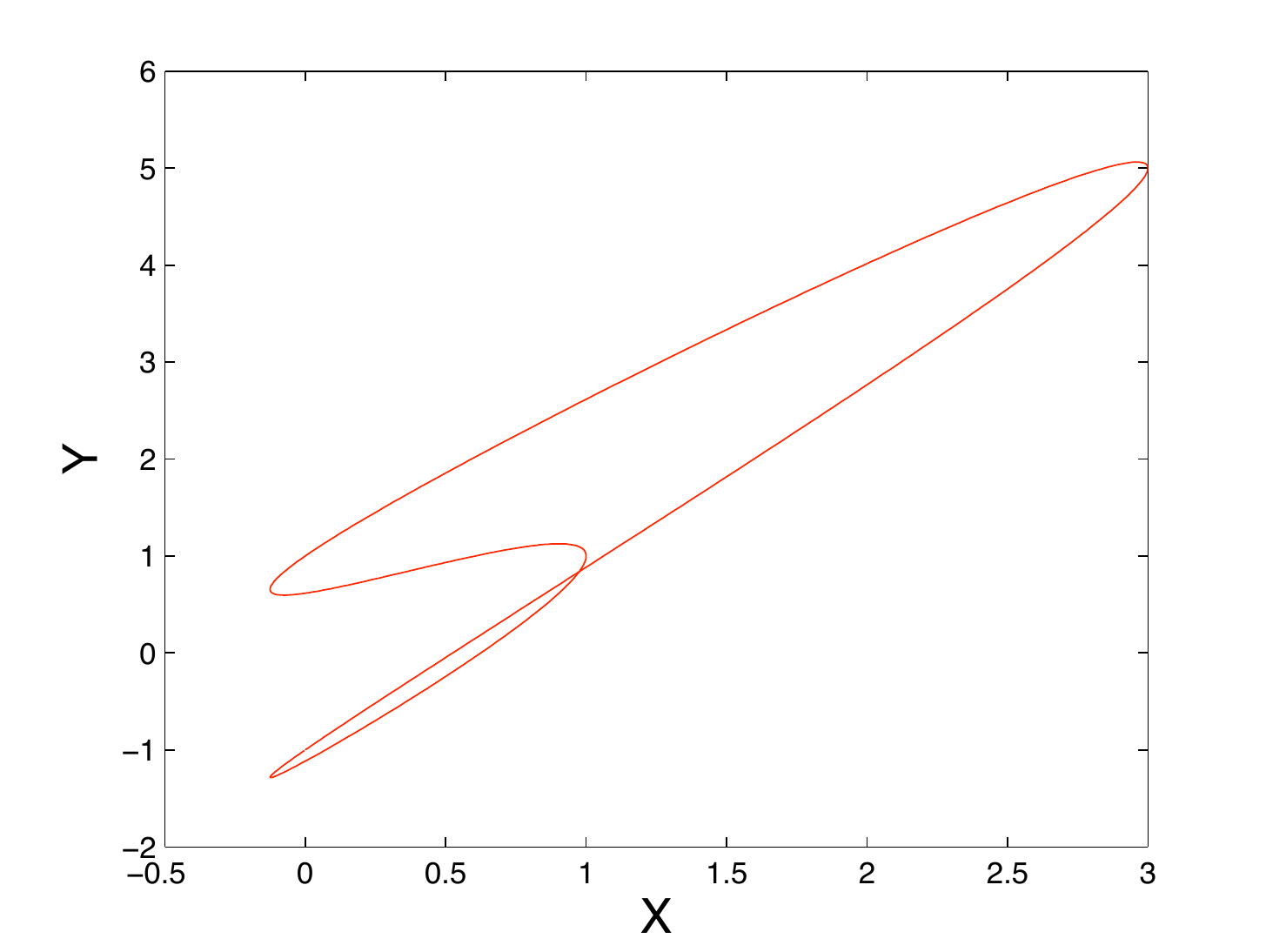,height=1.5in}\\
(a) curve $\gamma(t)$&(b)  special affine transformation $\br\gamma(t)$ &full affine transformation $\br{\gamma_a(t)}$
\end{tabular}}
 \label{fig:2dcurves}
\end{figure}

The integral invariants derived in the previous section depend on
the choice of the initial point and the parameterization of a curve.
For instance, consider a planar curve $\gamma(t)=(1/2\,\sin t -\cos
t +1, \sin ^2 t+\cos t -1),$  shown in Figure~\ref{fig:2dcurves}-a.
A curve $\br\gamma(t)$ (Figure~\ref{fig:2dcurves}-b) is obtained
from  $\gamma(t)$ by a special affine transformation $\left(
                                 \begin{array}{cc}
                                   2 & 1 \\
                                   2 & 1.5 \\
                                 \end{array}
                               \right)
$. A curve $\br{\gamma_a(t)}$ (Figure~\ref{fig:2dcurves}-c) is
obtained from  $\gamma(t)$ by a full affine transformation $\left(
                                 \begin{array}{cc}
                                   2 & 2 \\
                                  4 & 3 \\
                                 \end{array}
                               \right)
$. 

\begin{figure}[h]
\centerline{
\begin{tabular}{cc}
\epsfig{figure=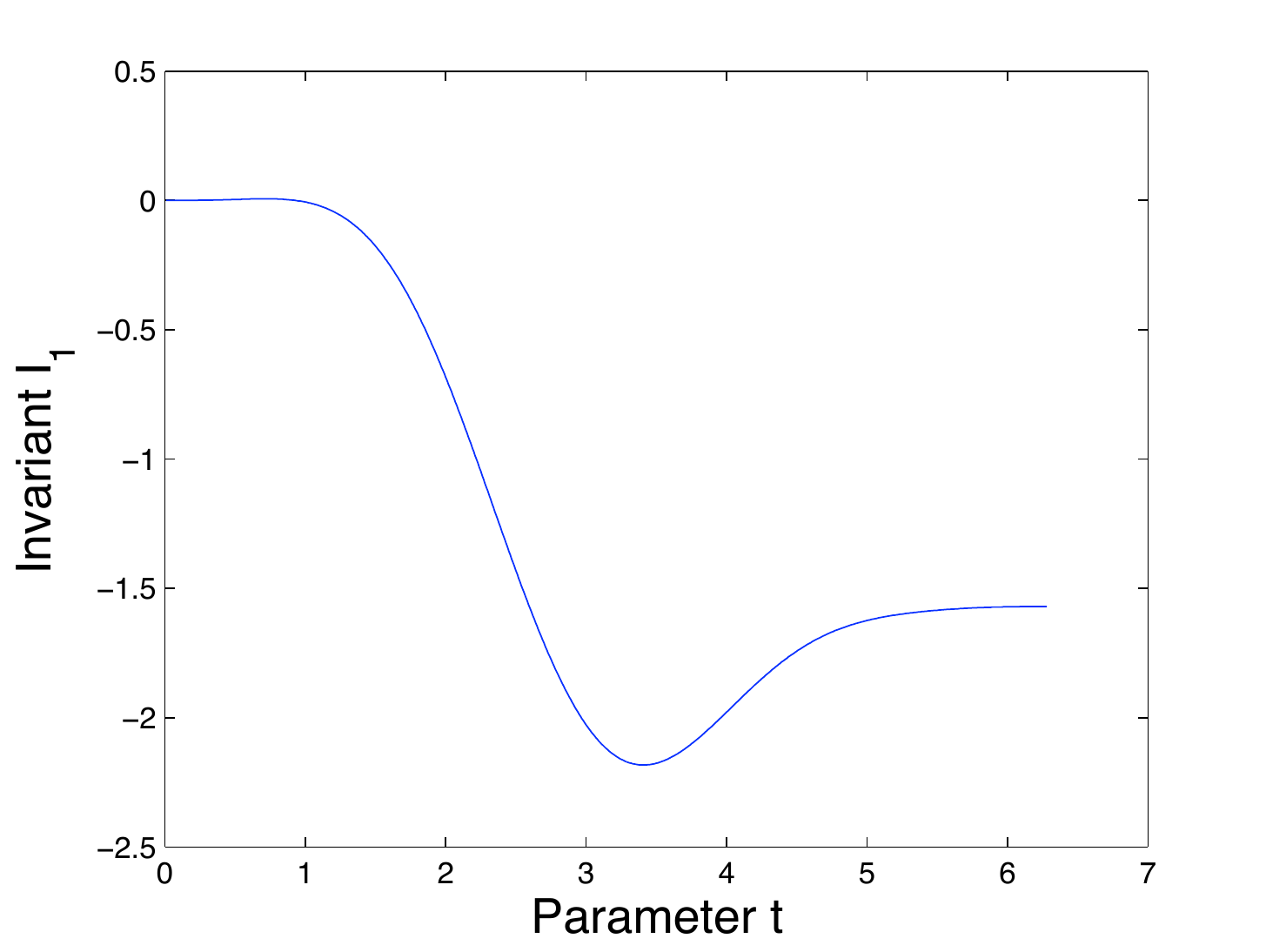,height=2in} &
\epsfig{figure=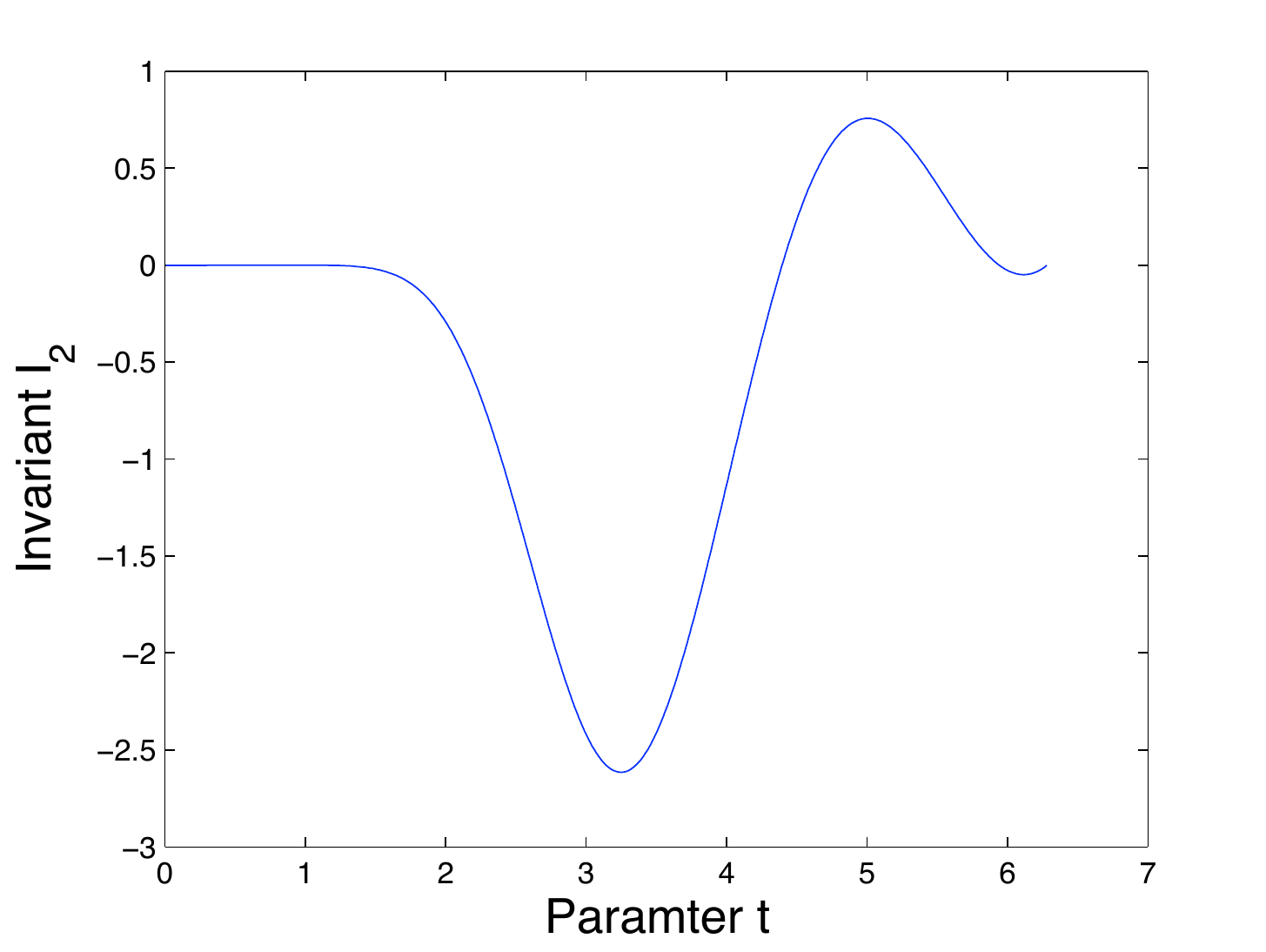,height=2in}  \\
(a) Invariant $I_1$ for  $\gamma(t)$ and
$\br{\gamma(t)}$&(b) Invariant $I_2$ for  $\gamma(t)$
and $\br{\gamma(t)}$\\
\epsfig{figure=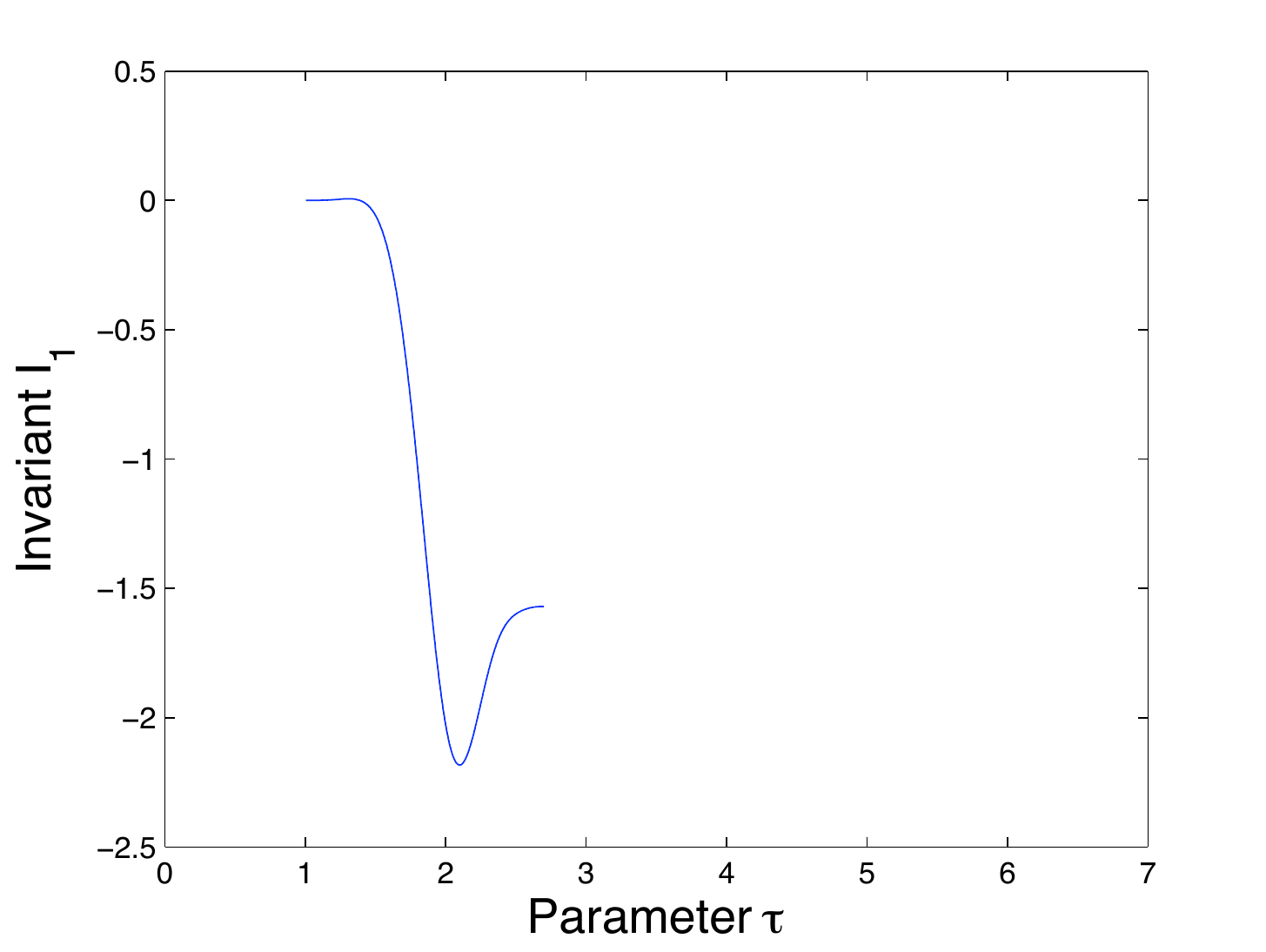,height=2in} &
\epsfig{figure=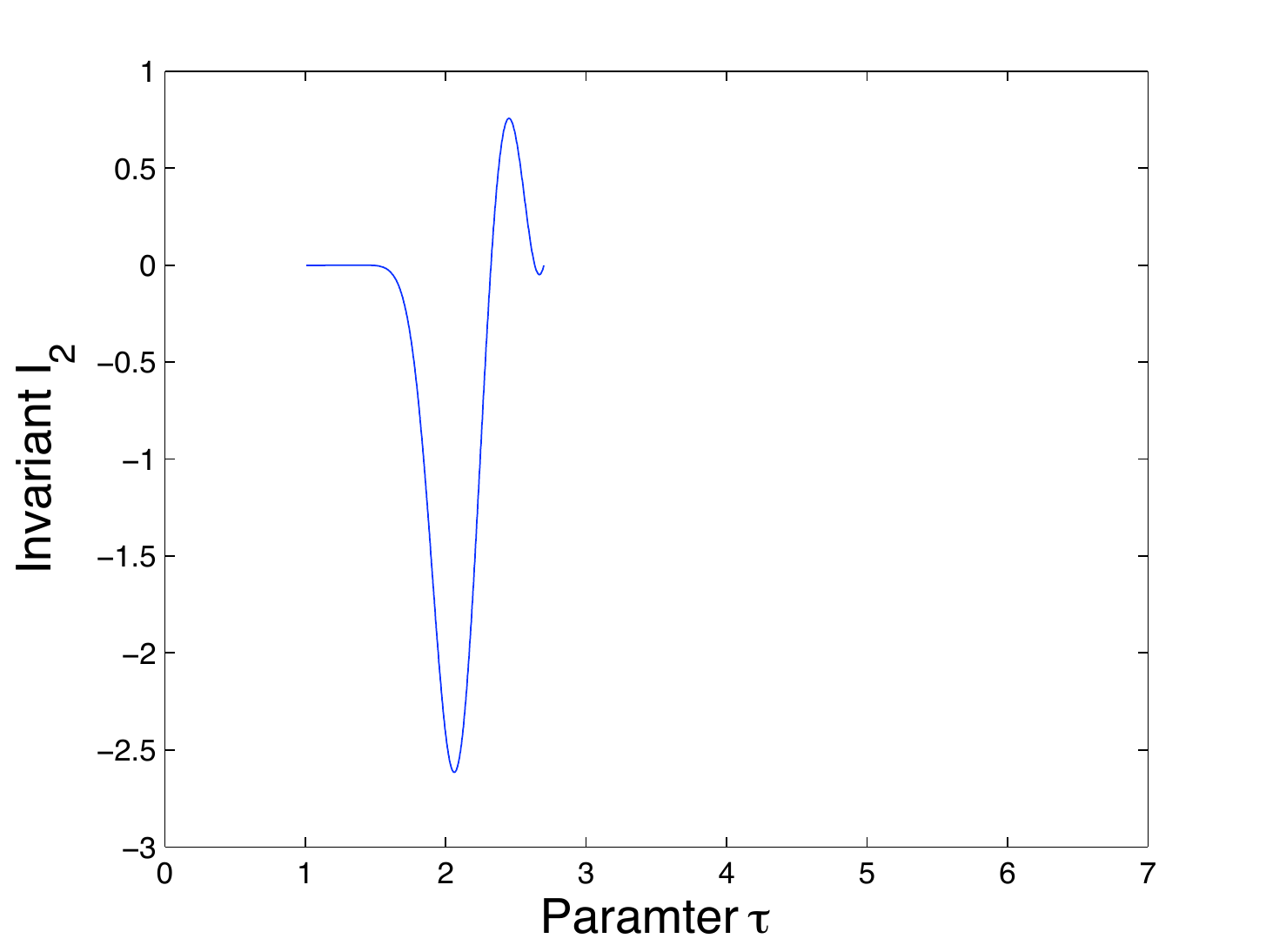,height=2in}  \\
(c) Invariant $I_1$ for
$\gamma(\tau)$ and $\br{\gamma(\tau)}$&(d) Invariant
$I_2$ for $\gamma(\tau)$ and $\br{\gamma(\tau)}$
\end{tabular}}
\caption{Dependence of invariants on re-prametrization: $\tau=\sqrt{t+1}$}
\label{fig:I1_I2}
\end{figure}

The integral invariants $I_1$ and $I_2$ for curves in
Fig.~\ref{fig:2dcurves}-a and Fig.~\ref{fig:2dcurves}-b with a
matching parameterization coincide and are shown in
Figure~\ref{fig:I1_I2}-a and Figure~\ref{fig:I1_I2}-b.

As illustrated  in Figure~\ref{fig:I1_I2}-c and
Figure~\ref{fig:I1_I2}-d, these invariants change under
reparametrization $\tau=\sqrt{t+1}$. Therefore the graph of
invariants with respect to an arbitrary parameter can not be used
for curves comparison. In theory one can achieve a uniform affine
invariant curve parameterization by using an affine analog of the
Euclidean arc-length parameter $d\alpha=\kappa^{1/3}ds$, where
$\kappa$ is Euclidean curvature and $ds$ is Euclidean arc-length. We
would like however to keep our methods derivative free. Even when
the uniform parameterization  is achieved, the dependence of the
invariants on the choice  of the initial point presents  another
comparison challenge for matching closed curves, or for matching
parts of the contours.

The signature construction, proposed in this section, leads to
classification methods which are independent of parameterization and
of the initial point.   Inspired by  signatures based on
differential invariants \cite{COSTH98}, we use integral invariants
to construct two types of signatures that classify curves under
affine transformation: the global signature  and the local
signature. {\it Global integral signature} is independent of
parametrization, but \emph{is dependent} on the choice of the initial
point and can not be used to compare partial contours. {\it Local
integral signature} is independent of both the initial point and
parametrization. They can be used to compare parts of the contours
and therefore can be used on images with occlusions. As our
experiments illustrate they are slightly more sensitive  to noise
than global signatures, but still provide robust classification
results.

\subsection{Global Integral Affine Signature}
A global integral signature of a curve is the variation of  one independent integral
invariant, evaluated on the curve,  relative to  another. If a
curve is mapped to another curve by  a group transformation, their
signatures coincide independently of the selected
parametrization. The global signature, however, does depend on a choice of the initial point.
\subsubsection{ Global affine signature for curves in 2D}
\begin{figure}[h]
\centerline{
\begin{tabular}{cc}
\epsfig{figure=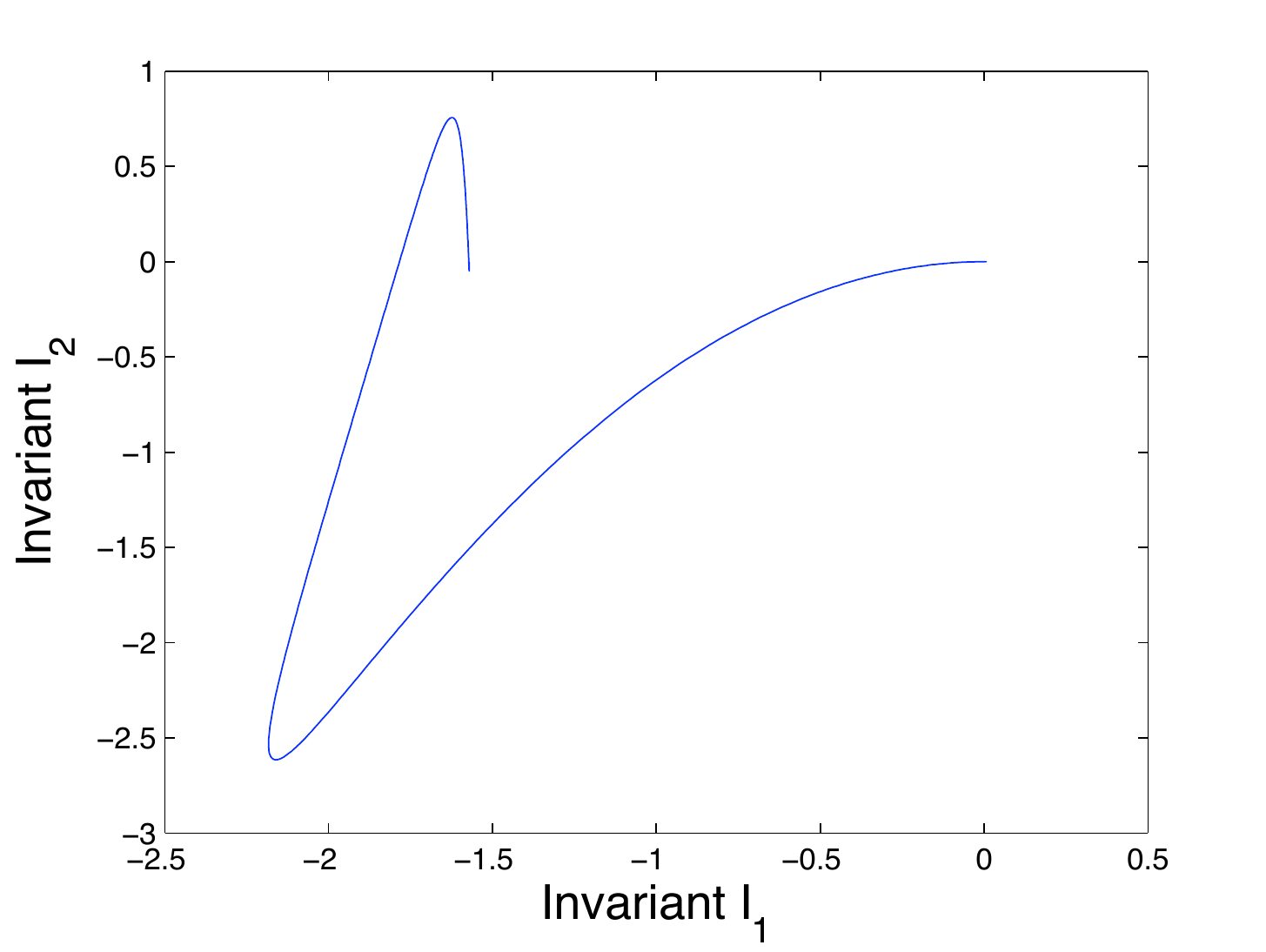,height=2in}
\end{tabular}}
\caption{ Signatures for  $\gamma(t)$,
$\br{\gamma(t)}$,$\gamma(\tau)$, and $\br{\gamma(\tau)}$ coincide \hskip7mm
} \label{fig:2dsign}
\end{figure}

The special affine signature of a plane curve $\gamma(t)$ is
constructed by, first, evaluating invariants $I_1$ and $I_2$ in
Eq.\Eq{SL2inv} on this curve, and then plotting the parameterized
curve $\left(I_1(t),I_2(t)\right)$ in $\RR^2$. For instance, the
signature of the planar curve   $\gamma(t)$ shown on
Figure~\ref{fig:2dcurves}-a is a plane curve in
Figure~\ref{fig:2dsign}. Moreover, the signature of the curve
$\br{\gamma(t)}$, related to $\gamma(t)$ by an affine transformation
(Figure~\ref{fig:2dcurves}-b), as well as their reparametrization
$\gamma(\tau)$ and $\br{\gamma(\tau)}$ coincide with the signature
of $\gamma(t)$.
\begin{figure}[h]
\centerline{
\begin{tabular}{cc}
\epsfig{figure=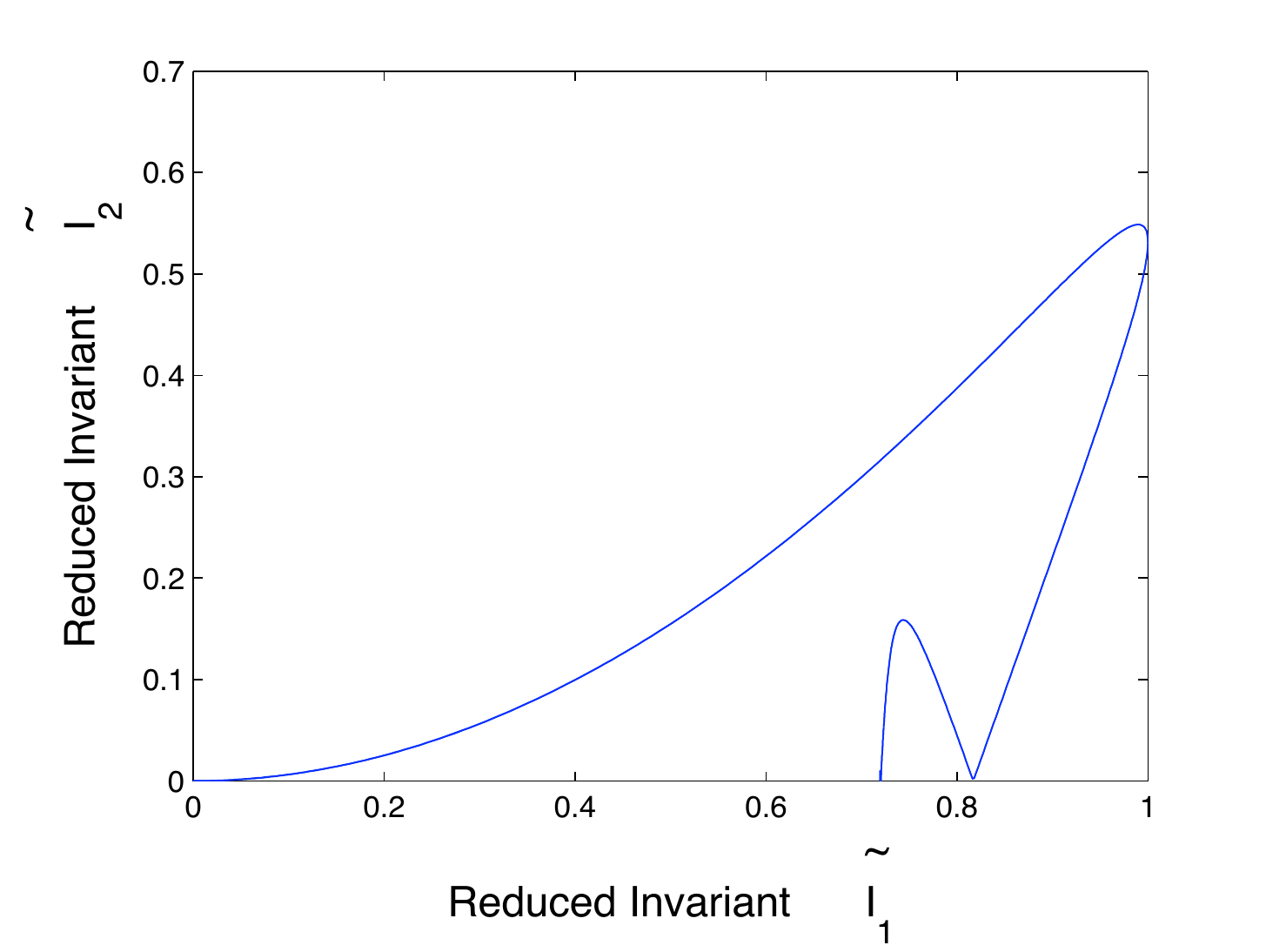,height=2in}
\end{tabular}}
\caption{Full affine  signatures
 for curves $\br{\gamma(t)}$ and $\br{\gamma_a(t)}$ coincide}
  \label{fig:fullsig2d}
\end{figure}

Similarly, a full affine signature can be defined  as a
parameterized plane curve $\left(I_2^\A,I_3^\A\right)$ defined by
invariants in Eq.\Eq{GL2inv}. Alternatively we use the two special
affine invariants to cancel the effects of reflections and arbitrary
scalings, \eq{ninv2d}{ \tilde
{I_{1}}(t)=\frac{\left|I_{1}(t)\right|}{\max_t\left|I_{1}\right|},\quad
\tilde{I_{2}}(t)=\frac{\left|I_{2}\right|}{\max_t(I_{1}^2)}.} Both
invariants are reduced relative to the range of
$\left|I_{1}\right|$. It is not difficult to show that
${\max_t\left|I_{1}\right|}=0$  on $\gamma$ if and only if $\gamma$
is a straight line. In this case the affine signature does not
exist, but straight line regions can be easily detected by other
means.   The range of $\tilde {I_1}$ is from 0 to 1. The full affine
signature of a plane curve $\gamma(t)$ is obtained by, first,
evaluating $\tilde{I_1}$ and $\tilde{I_2}$ on this curve and then by
plotting the parameterized curve
$\left(\tilde{I_1}(t),\tilde{I_2}(t)\right)$ in $\RR^2$. For
example, curves in Fig.~\ref{fig:2dcurves}-a and
Fig.~\ref{fig:2dcurves}-c are related by a non-area-preserving
affine transformation. Their   full affine signatures coincide as
shown on  Figure~\ref{fig:fullsig2d}.

\subsubsection{ Global Affine Signatures for Curves in 3D}
To construct special affine signatures for spatial curve we use
invariants $J_1$ and $J_2$   given by Eq.\Eq{3dinv}.  Similarly to
2D case, the special affine signature of a spatial curve $\beta(t)$
is obtained by, first, evaluating $J_1$ and $J_2$ on this curve, and
then  plotting the parameterized curve $\left(J_1(t),J_2(t)\right)$
in $\RR^2$.

\begin{figure}[h]\caption{}
\centerline{
\begin{tabular}{ccc}
\epsfig{figure=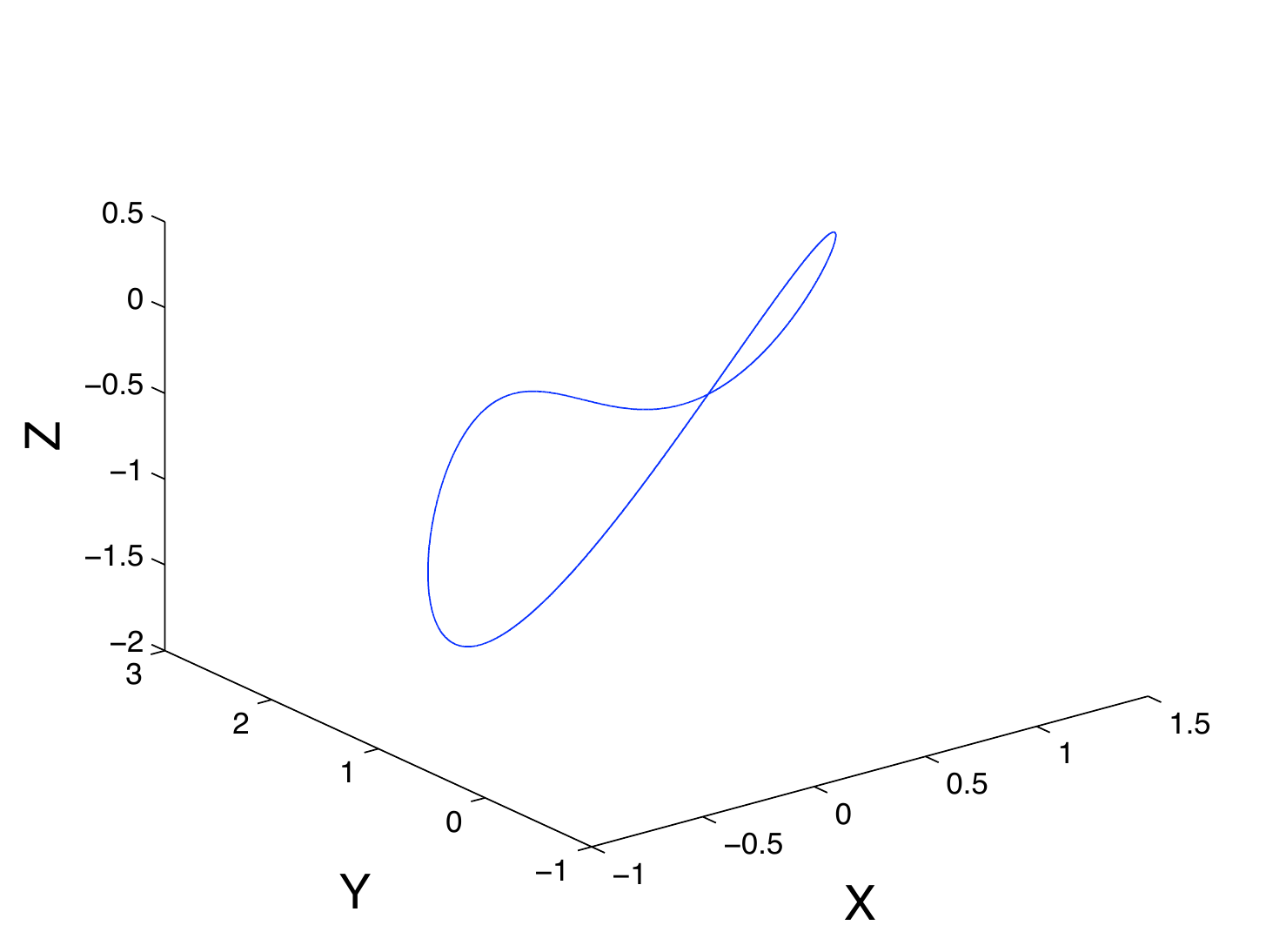,height=1.5in} &
\epsfig{figure=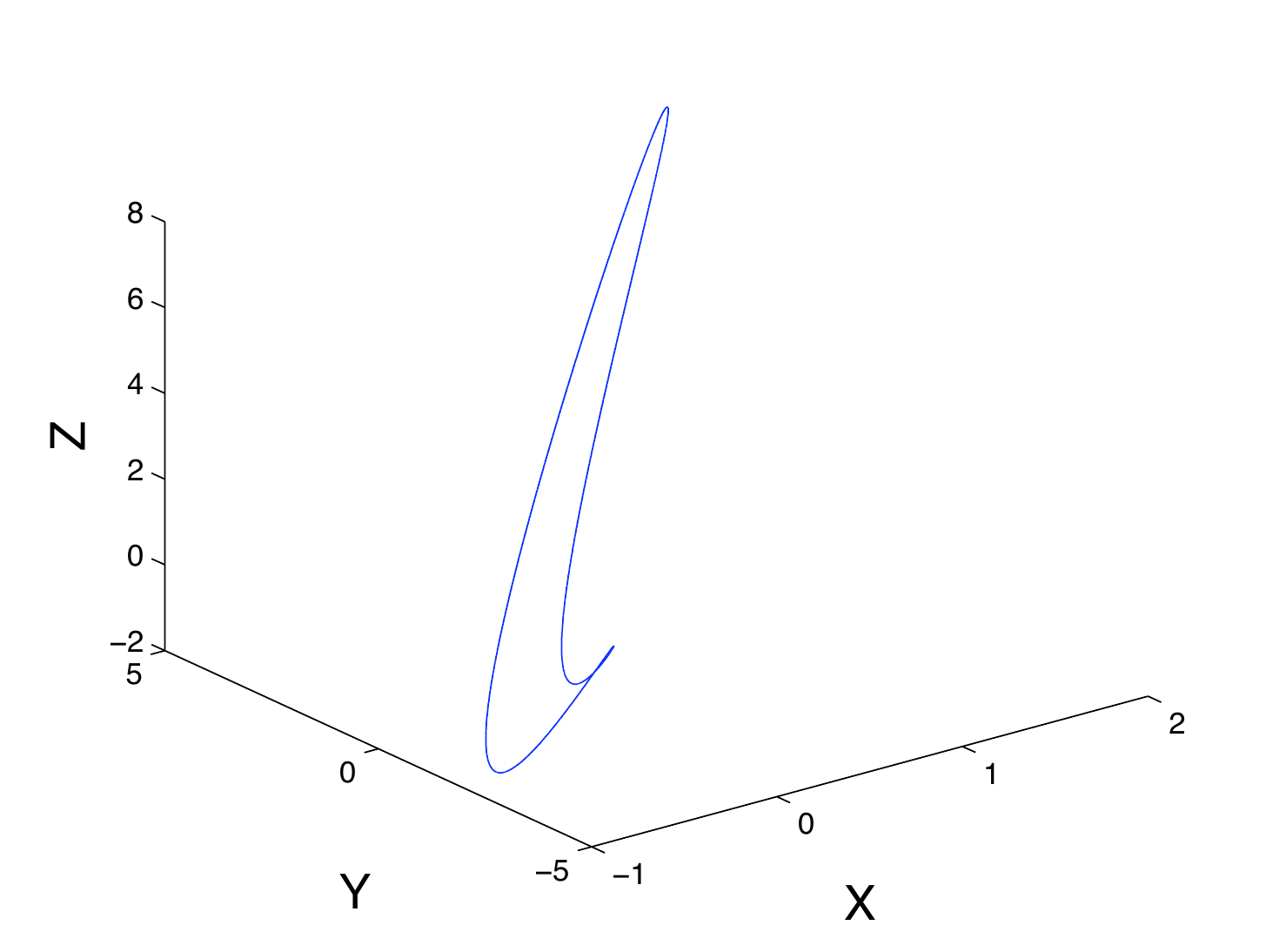,height=1.5in} &
 \epsfig{figure=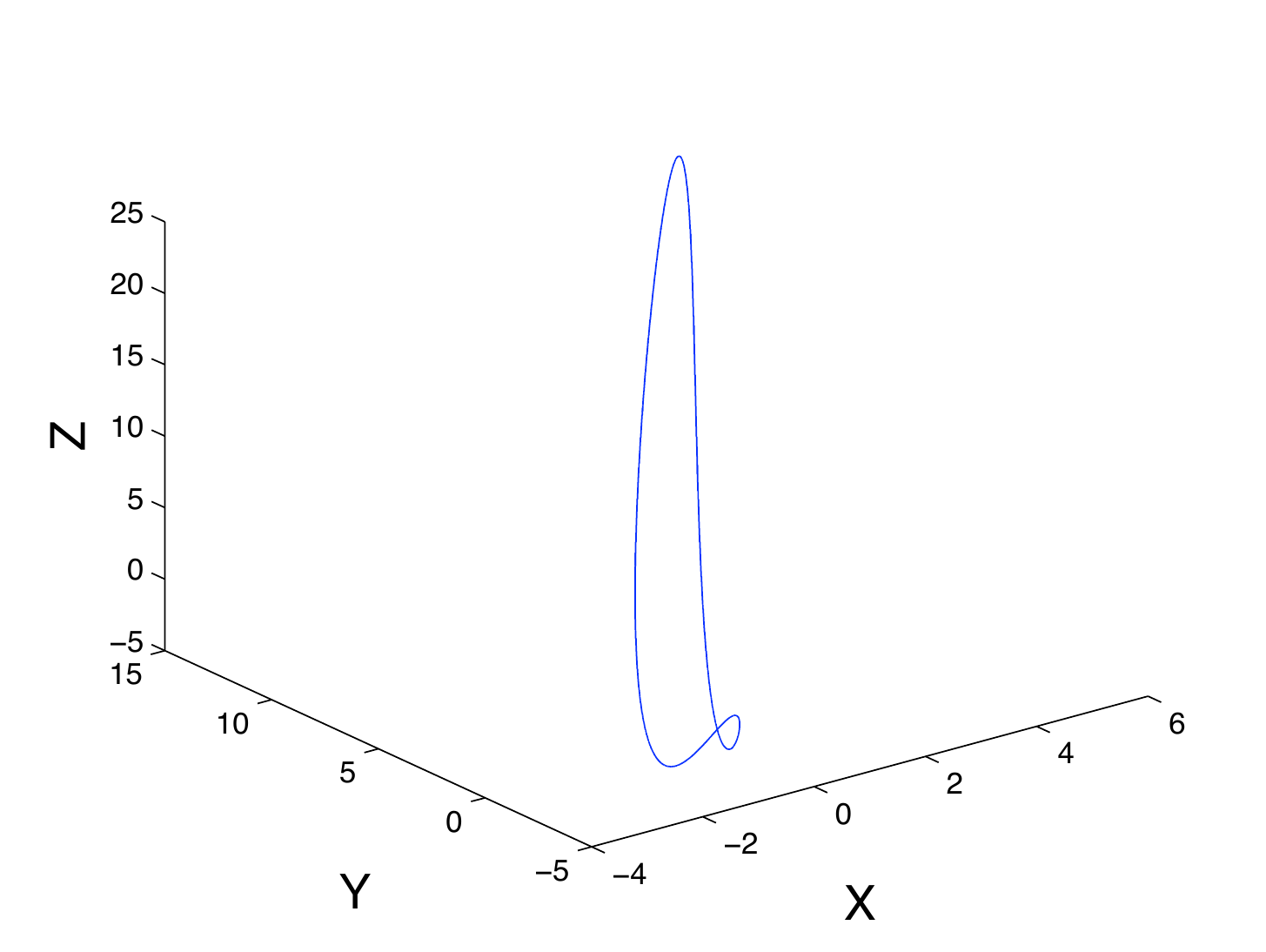,height=1.5in}\\
(a) original curve $\beta(t)$&(b) special affine
transformation $\br{\beta(t)}$&(c) full affine
transformation $\br{\beta_a(t)}$
\end{tabular}}
\label{fig:3dcurves}
\end{figure}
\begin{figure}[h]
\begin{center}
\epsfig{figure=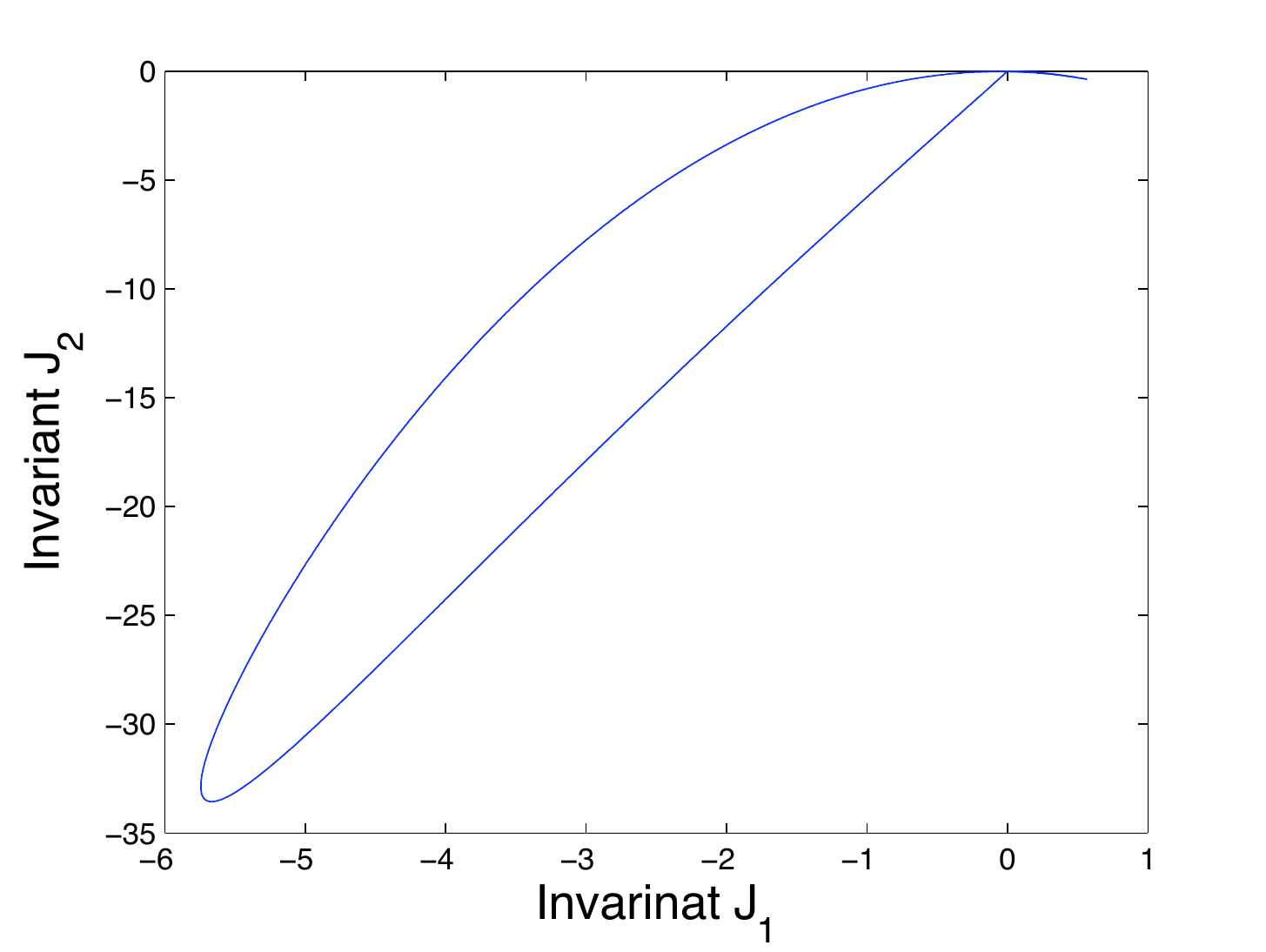,height=2in}
\caption{Signatures for  $\beta(t)$ and $\br{\beta(t)}$
coincide.}\label{fig:3dsign}
\end{center}
\end{figure}

For example, the signature of a  spatial curve $\beta(t)=(\sin t
-1/5\,  \cos ^2 t +1/5,\, 1/2\,\sin t -\cos t +1, \sin ^2 t+\cos t
-1),$ shown in Figure~\ref{fig:3dcurves}-a, is the plane curve shown
on Fig~\ref{fig:3dsign}. A curve $\br{\beta(t)}$ is obtained from
$\beta$ by a special affine transformation $\left(
                                                   \begin{array}{ccc}
                                                    0.3816 &   0.7631 &   1.1447 \\
                                                     1.9079  &  1.5263 &   2.2894\\
                                                      2.6710  &  3.0526  &  3.4341 \\
                                                   \end{array}
                                                 \right)
$. A curve $\br{\beta_a(t)}$ is obtained from $\beta$ by a full
affine transformation $\left(
                                                   \begin{array}{ccc}
                                                   1 &   2 &   3\\
                                                     4 &  5 &   6\\
                                                     9  &  8  &  7 \\
                                                   \end{array}
                                                 \right)
$. As Fig~\ref{fig:3dsign} illustrates, the special affine
signatures of $\beta(t)$ and $\br{\beta(t)}$ coincide.

\begin{figure}[h]
\begin{center}
\epsfig{figure=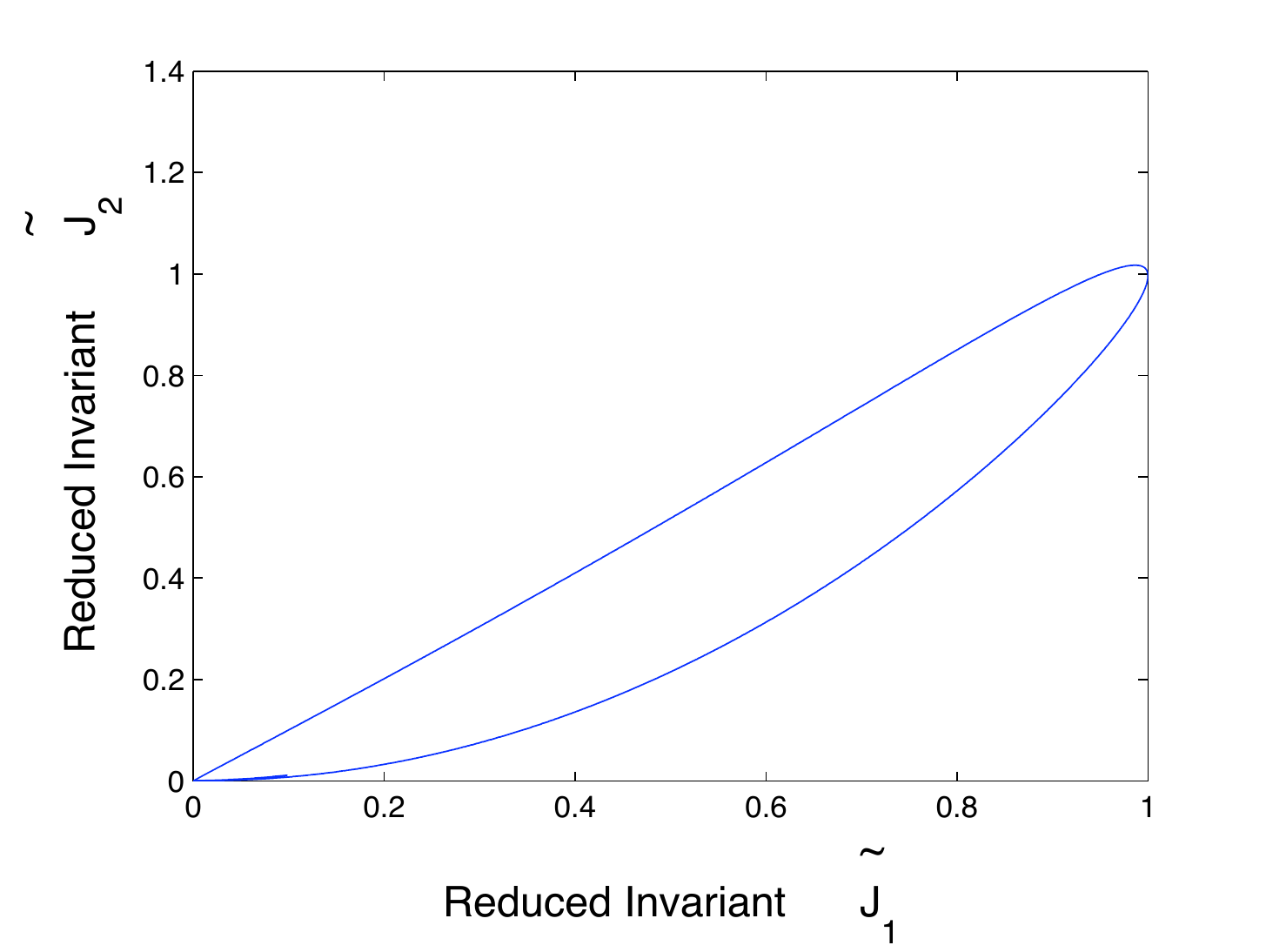,height=2in}
\caption{Full affine signatures for  $\beta(t)$, $\br{\beta(t)}$,
$\br{\beta_a(t)}$ }\label{fig:3dsignaffine}
\end{center}
\end{figure}
Similar to the 2D case, the full affine signature for curves in 3D is obtained by
reducing  special affine invariants $J_1$ and $J_2$ by the range of
$\left|J_1\right|$. \eq{A3inv}{
\tilde{J_{1}}(t)=\frac{\left|J_{1}(t)\right|}{\max_t\left|J_{1}\right|},\quad
\tilde{J_{2}}(t)=\frac{J_{2}(t)}{\max_t(J_{1}^2)}.}

The full affine signature of a spatial curve $\beta(t)$ is obtained
by first evaluating $\tilde{J_1}$ and $\tilde{J_2}$ on this curve,
and by then plotting the parameterized curve
$\left(\tilde{J_1}(t),\tilde{J_2}(t)\right)$ in $\RR^2$. The full
affine signatures of $\beta$, $\bar\beta$ and $\bar\beta_a$ coincide
as shown in Fig~\ref{fig:3dsignaffine}.

The advantage of global signatures is their independence of
parametrization, whereas  the result of evaluation of invariants
$J_1$ and $J_2$ on a curve depends on the choice of parametrizations
similarly to  $I_1$ and $I_2$ in 2D case. The disadvantage of global
signatures is in their dependence on the choice of the initial point
of a curve. The local signature construction in the next section
overcomes this dependence.

\subsection{Local Integral Affine Signatures}
\begin{figure}[h]
\centerline{
\epsfig{figure=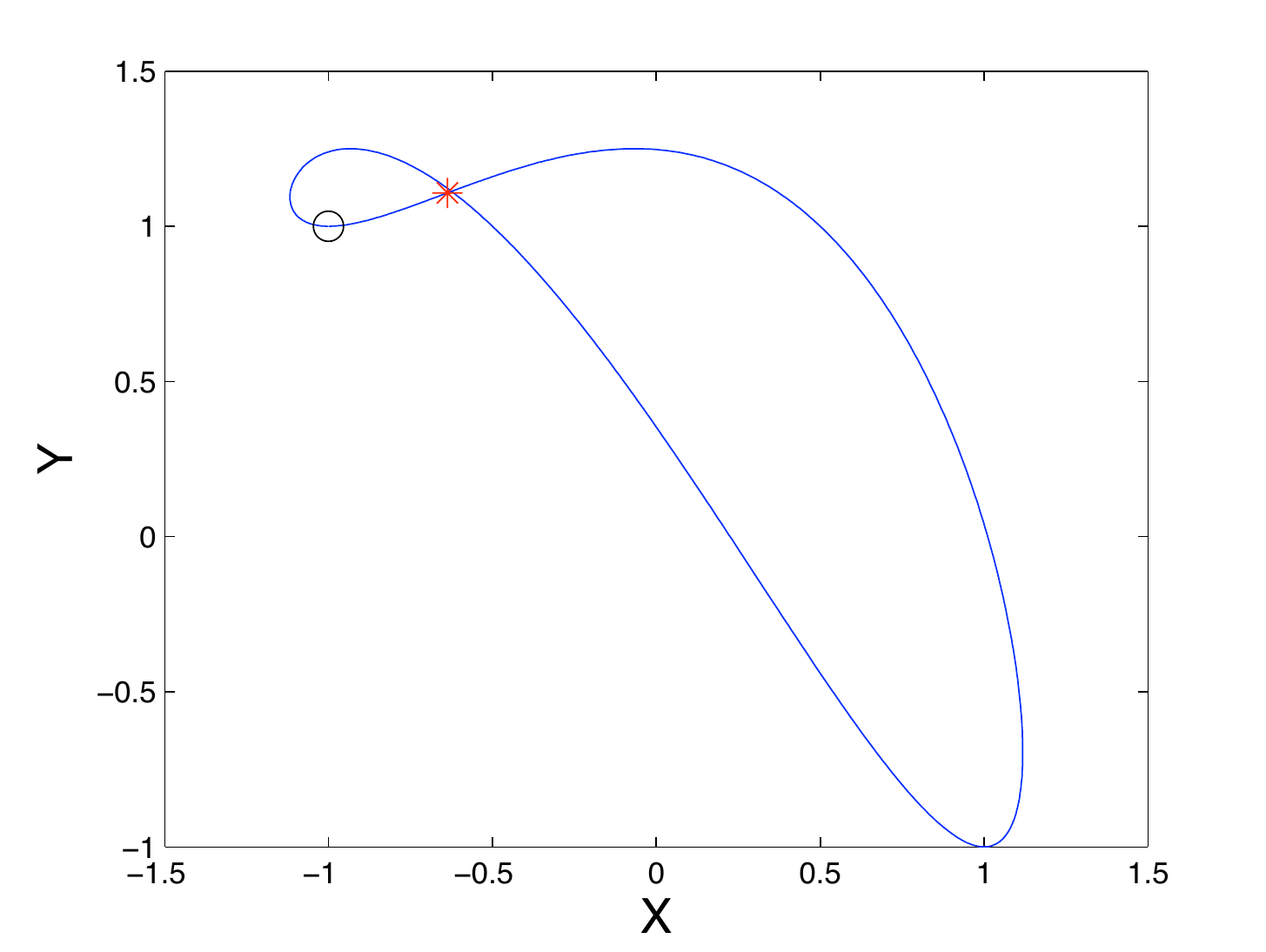,height=2in}} \caption{A planar
curve with two different choices of the initial points} \label{fig:2d_2initials}
\end{figure}
\begin{figure}[h]
\centerline{
\begin{tabular}{cc}
\epsfig{figure=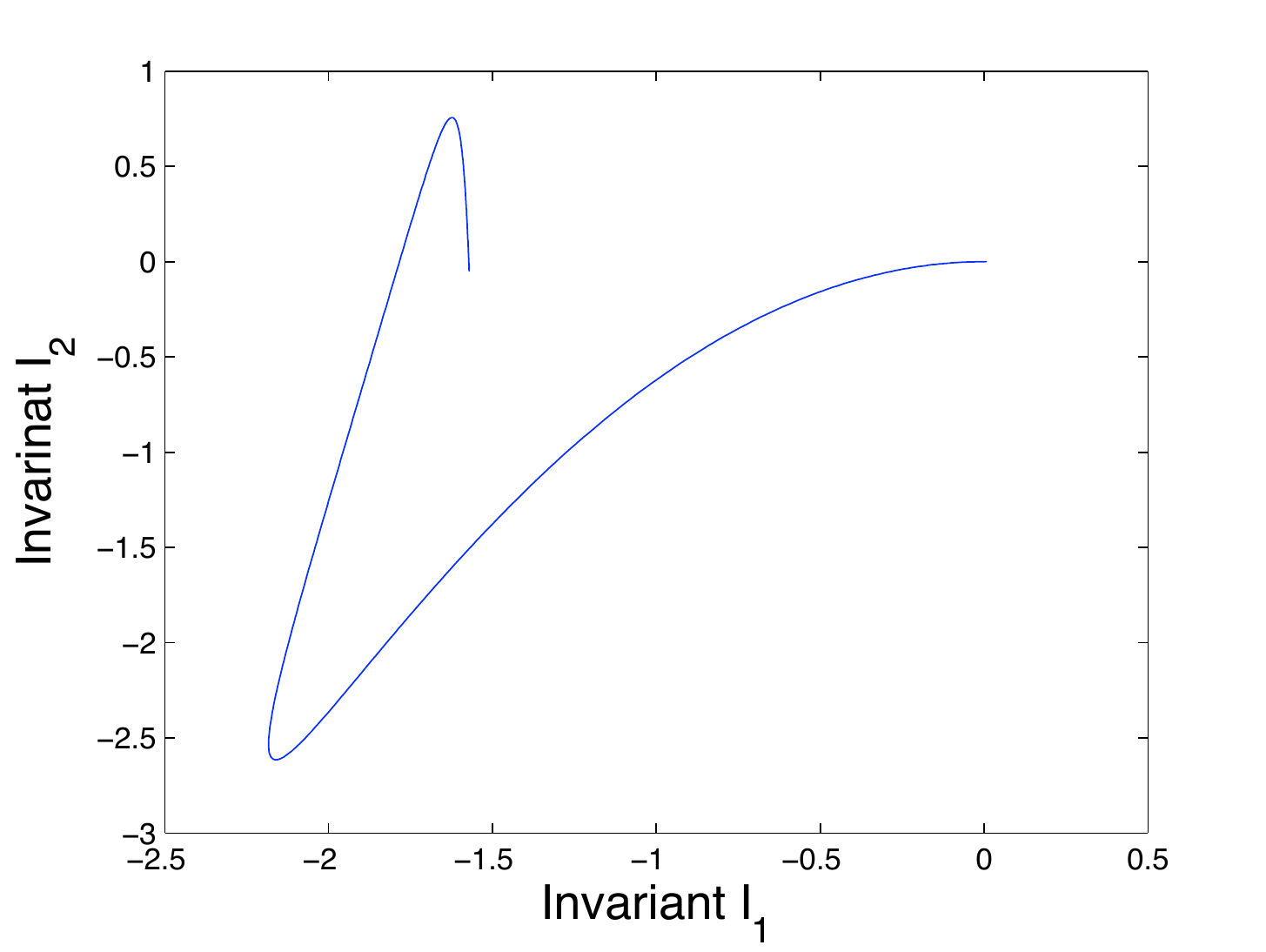,height=2in} &
\epsfig{figure=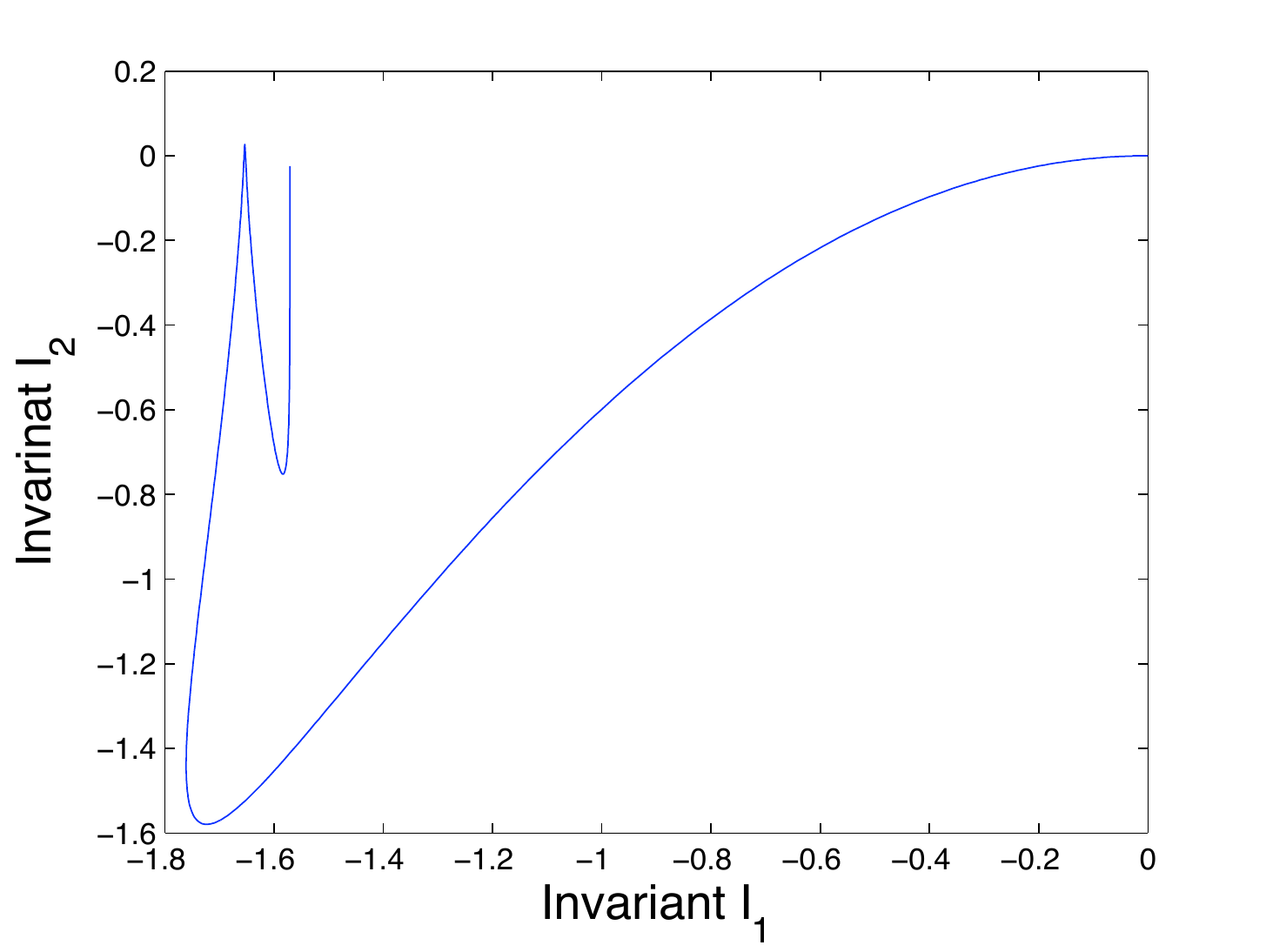,height=2in} \\
(a) Initial point is the black circle in Fig.\ref{fig:2d_2initials}
&(b) Initial point is the red star in Fig.\ref{fig:2d_2initials}
\end{tabular}}
\caption{Global signatures for the same curve with two different choices of the initial point}\label{fig:2d_sig_2ini}
\end{figure}

The signatures defined in the previous section can not be used for
the classification unless the initial point of a curve is known.
This becomes an obstacle for comparing closed curves or for matching
parts of contours. For illustration, let us choose two different
initial points,  black circle or red star, on the planar  curve in
Figure~\ref{fig:2d_2initials}. The resulting global affine
signatures are  different as illustrated in
Figures~\ref{fig:2d_sig_2ini}-a and Figures~\ref{fig:2d_sig_2ini}-b.
We overcome the dependence on the initial point by introducing local
signatures. To proceed with the construction of the local signature,
 we replace the integration from the initial point with integration
on local segments. To retain affine invariant properties of the
signatures, we need to partition the curve in an invariant manner.
Such partition can be  achieved using the notion of affine
arc-length from classical differential geometry. Our goal, however,
is to  propose a derivative free method, and so we use the lowest
order integral invariants, namely $I_1$ for plane curves and $J_1$
for spatial curve to obtain an equi-affine  partition of a given
curve. The details are described in the following subsections.
\subsubsection{ Local  Affine Signatures for Curves in 2D}
We will use $I_1$ to partition  a given  curve into equi-affine
sub-segments. Assume that  $\gamma$ is parametrized by $t\in[0,1]$.
For this purpose we define an evaluation of  invariants on
sub-segments  of $\gamma$. Recall that the integration in the
integral variables is performed from the initial  point $\gamma(0)$
to a current point on the  curve $\gamma(t)$. For instance,
$I_1(t)=\int_0^t XdY-\frac 12 XY$, where $X=x(t)-x(0)$ and
$Y=y(t)-y(0)$. Thus $I_1(t)$ is a function from $[0,1]$ to $\R$.

We define the evaluation of an invariant on sub-segments  of
$\gamma$ by  treating  the  starting point of a segment as  the
initial point, and computing the value of integral variables at the
end point. In particular, for a sub-segment  defined by the
parameter range $[p,q]\subset[0,1]$, we may compute the localization
$I_1^{[p,q]}=\int_{p}^{q}\left(x(t)-x(p)\right)dy(t)-\frac 12
\left(x(q)-x(p)\right)\left(y(q)-y(p)\right)$,  and similarly for
invariants $I_2$ and $I_3$ defined by Eq.\Eq{SL2inv}. We note that
the evaluation of  an invariant on a segment is a real number.

We choose a sufficiently small $\Delta>0$ and define   an
equi-affine partition  $0=t_0<t_1<\dots<t_N=1$ of the curve
$\gamma(t),\,t\in[0,1]$ into  sub-segments  by
 the condition
$$ \left|I_1^{[t_{i-1},t_i]}\right|=\Delta .$$
In practice we choose $\Delta$ proportionally to the maximum of the
absolute value of $I_1$, i.e., ~we choose an integer $M$ and set
$\Delta=\frac{\max_t\left|I_t\right|}{M}$.  Note that the total
number $N$ of segments that we obtain in general differs from $M$.
The \emph{local discrete special affine signature} of $\gamma$ is
defined by an evaluation of $I_2$ and $I_3$ on the intervals
$[t_{i-1}, t_i],\, i=1..N$, that is a set of points with coordinates
$\left(I_2^{[t_{i-1}, t_i]},I_3^{[t_{i-1}, t_i]}\right)\, i=1..N$.
\begin{figure}[h]
\centerline{
\epsfig{figure=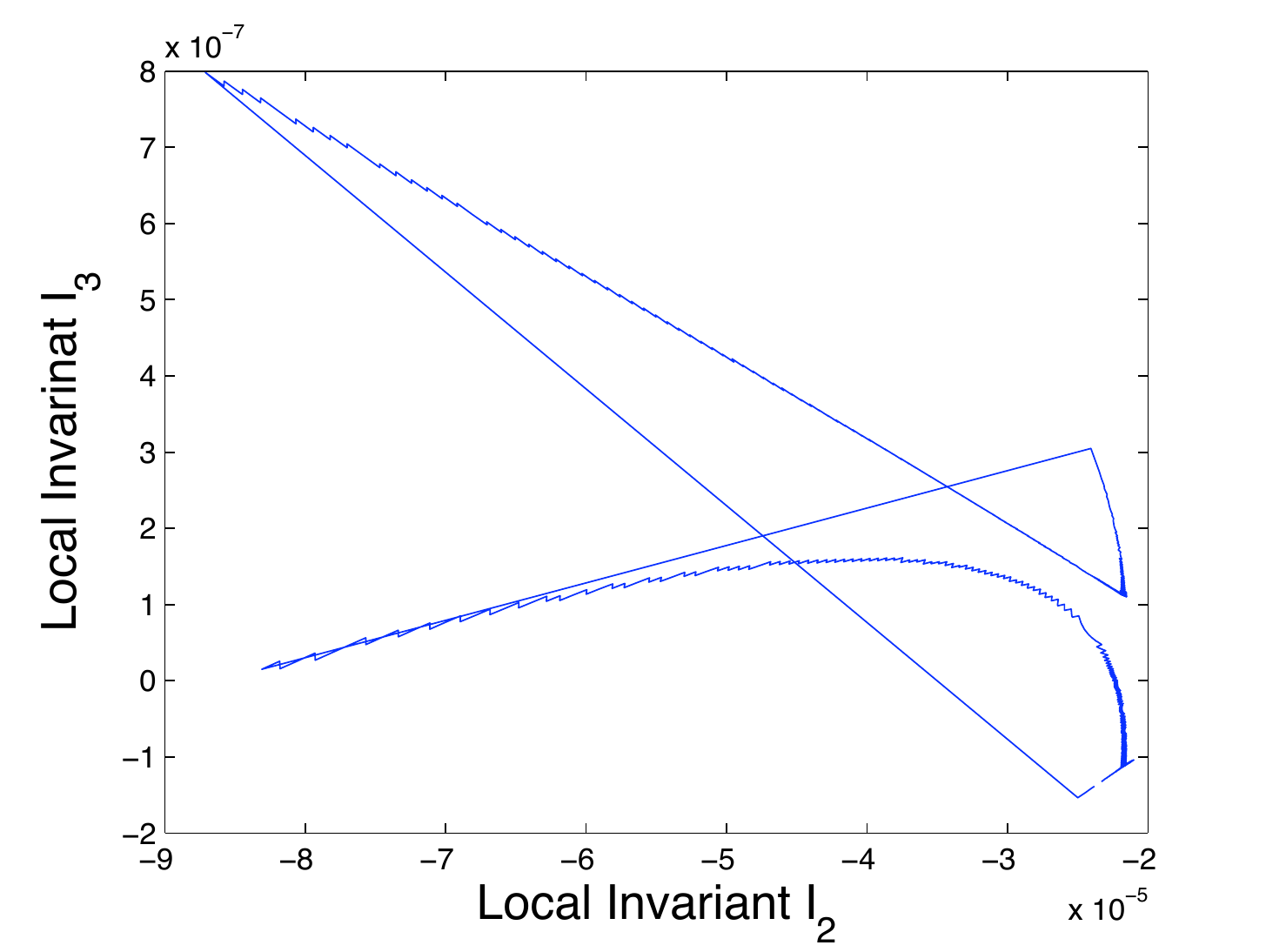,height=2in}}
\caption{Local special affine signature for the curve shown on Figure~\ref{fig:2d_2initials} }
\label{fig:2d_local_sig}
\end{figure}
Figure~\ref{fig:2d_local_sig} illustrates that the {local discrete
special affine signature}  for the curve shown in
Fig~\ref{fig:2d_2initials}, with different starting points
coincides. To obtain  \emph{local discrete affine signature} of
$\gamma$, we use reduced invariants, that is we plot $\left(\frac
{I_2^{[t_{i-1}, t_i]}}{max_t(I_1^2)}, \frac {I_3^{[t_{i-1},
t_i]}}{max_t\left|I_1^3\right|}\right),\, i=1..N$. \comment{discuss
the sensitivity to noise : finer partition may lead to higher
sensitivity}

\subsubsection{ Local  Affine Signatures for Curves in 3D}
For a spatial curve we proceed in a similar manner as for plane
curves. We use invariant  $J_1$ to partition a curve $\gamma(t),
t\in [0,1]$ into $N$  sub-intervals defined by
$a=t_0<t_1<\dots<t_N=b$ such that $ J_1^{[t_{i-1},t_i]}=\Delta,\,
i=1..N$, where $\Delta >0$ is proportional to the maximum of the
absolute value of $J_1$. We define a local special affine signature
by  evaluation of $J_2$ and $J_3$ on the intervals $[t_{i-1},
t_i],\, i=1..N$, that is by a  set of points on the plane with
coordinates $\left(J_2^{[t_{i-1}, t_i]},J_3^{[t_{i-1},
t_i]}\right)\, i=1..N$.

\begin{figure}[h]
\centerline{
\begin{tabular}{cc}
 \epsfig{figure=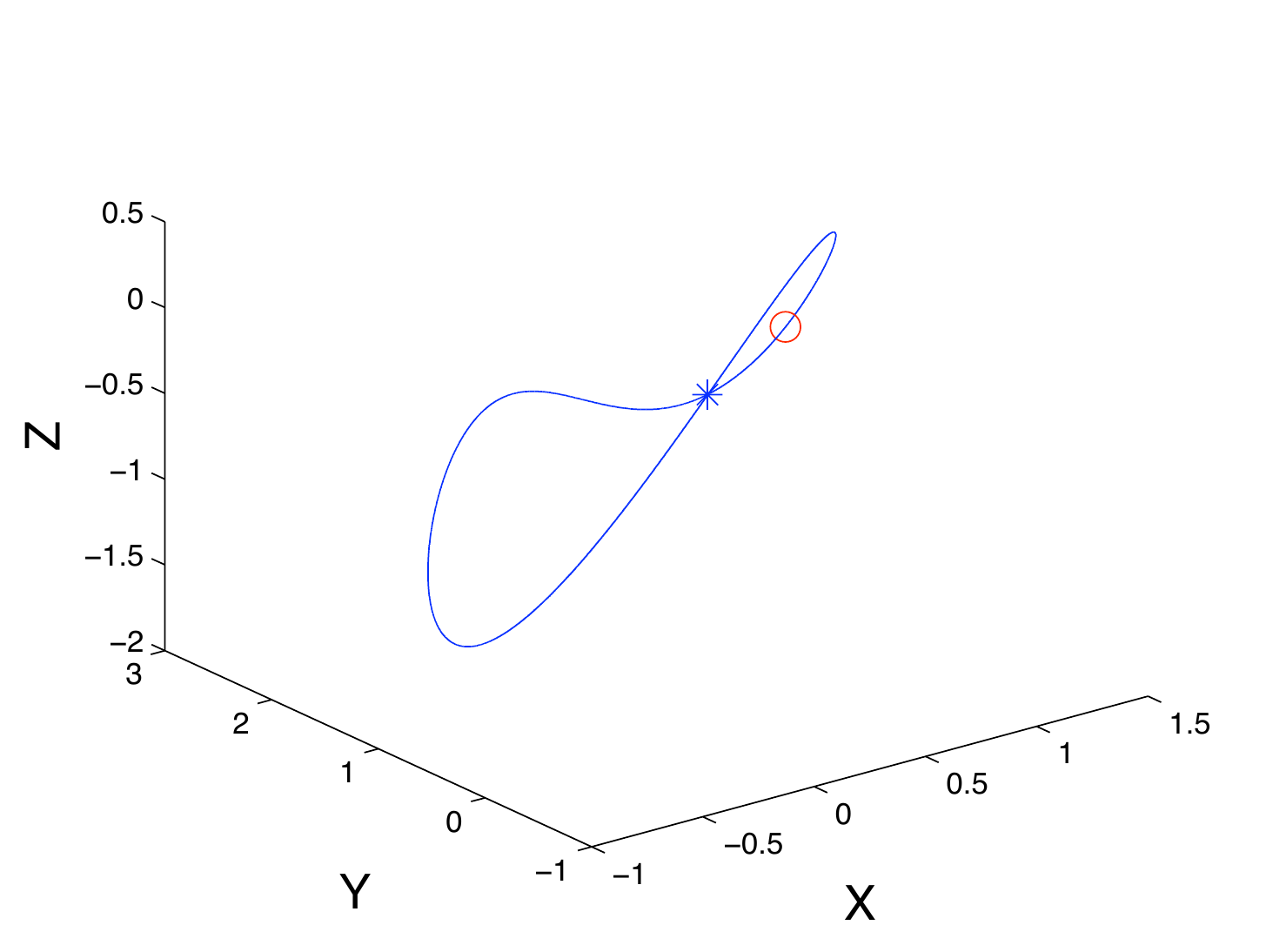,height=2in}
 &
\epsfig{figure=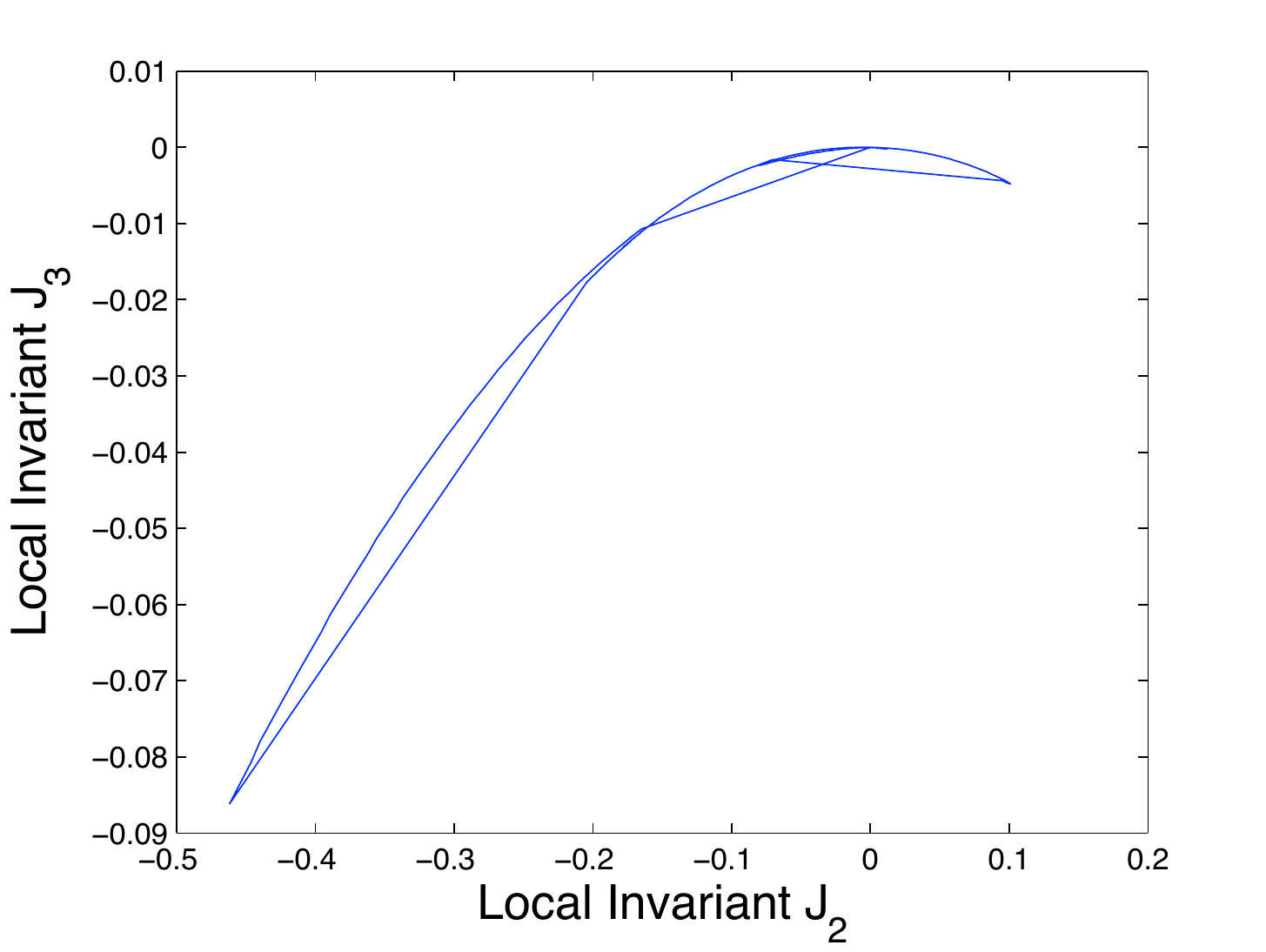,height=2in}
\\
(a) A curve with two different choices of an initial point &(b)
Local special affine signature
\end{tabular}}
\caption{A curve and its local special affine signature} \label{fig:3d_sig_2ini}
\end{figure}

 Figure ~\ref{fig:3d_sig_2ini}-b shows the local special affine signature for a curve  shown  on Figure ~\ref{fig:3d_sig_2ini}-a. The signature does not depend on our choice of initial point.

 \section{3D Object Classification}
 The features of many computer vision and pattern analysis problems
 are spatial curves. We can hence view the classification problem as that of
 assorting the similarity of  curves in 3D, in particular, when subjected to affine transformations.
 In this section, we apply integral special affine invariants
$J_{1}$ and $J_{2}$, the global special affine signature, and the
local special affine signature to classify curves in 3D under special affine transformations. The
performance of each of the proposed methods is evaluated. Applying
these invariants to classification of 3D objects based on a set of
characteristic spatial curves is in line of \cite{Aouada07icassp},
and will be considered in subsequent publications.
\subsection{Experimental Design}

\begin{figure}
\centering
   \includegraphics[height=2in]{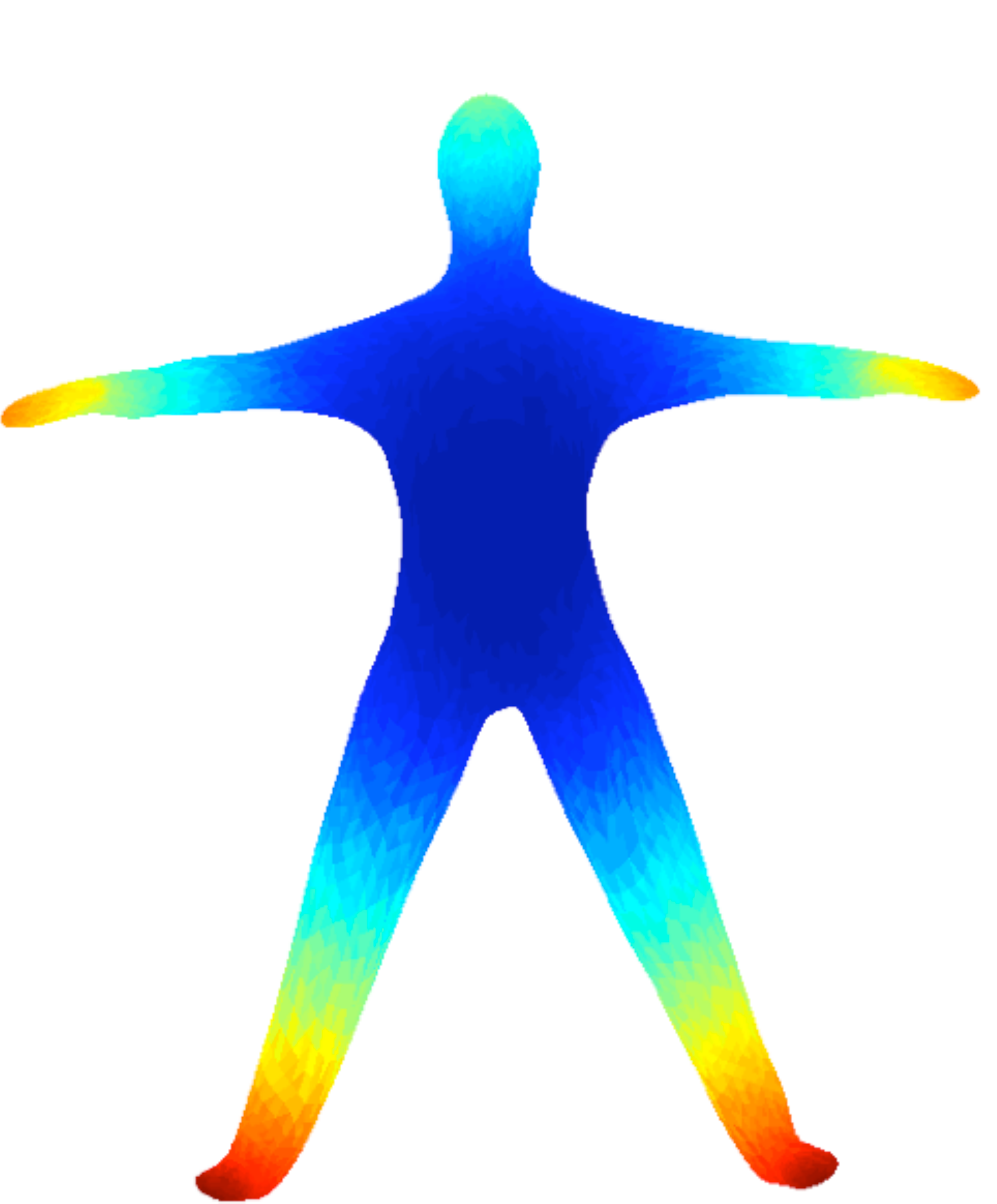}
   \includegraphics[height=2in]{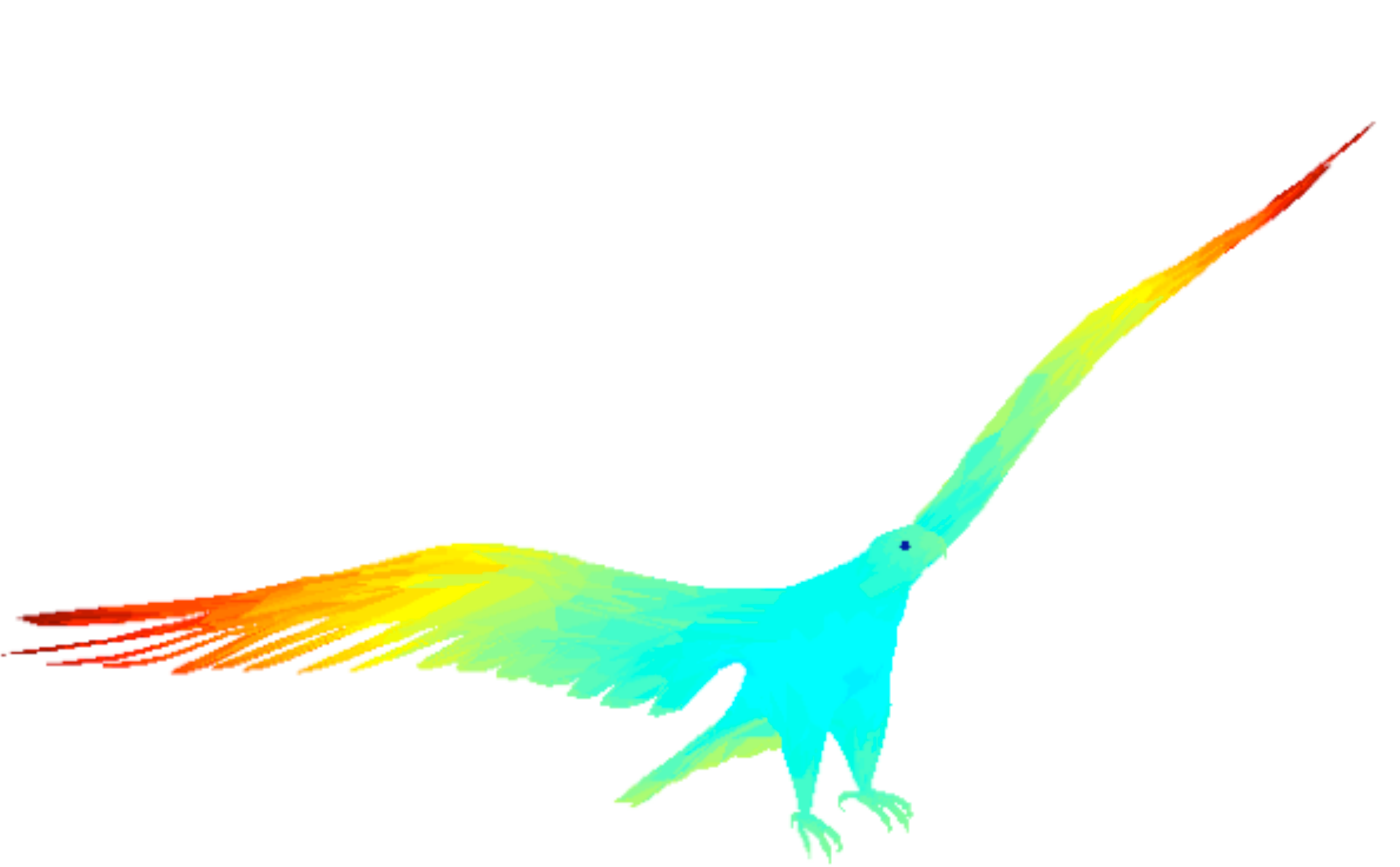}
   \includegraphics[height=2in]{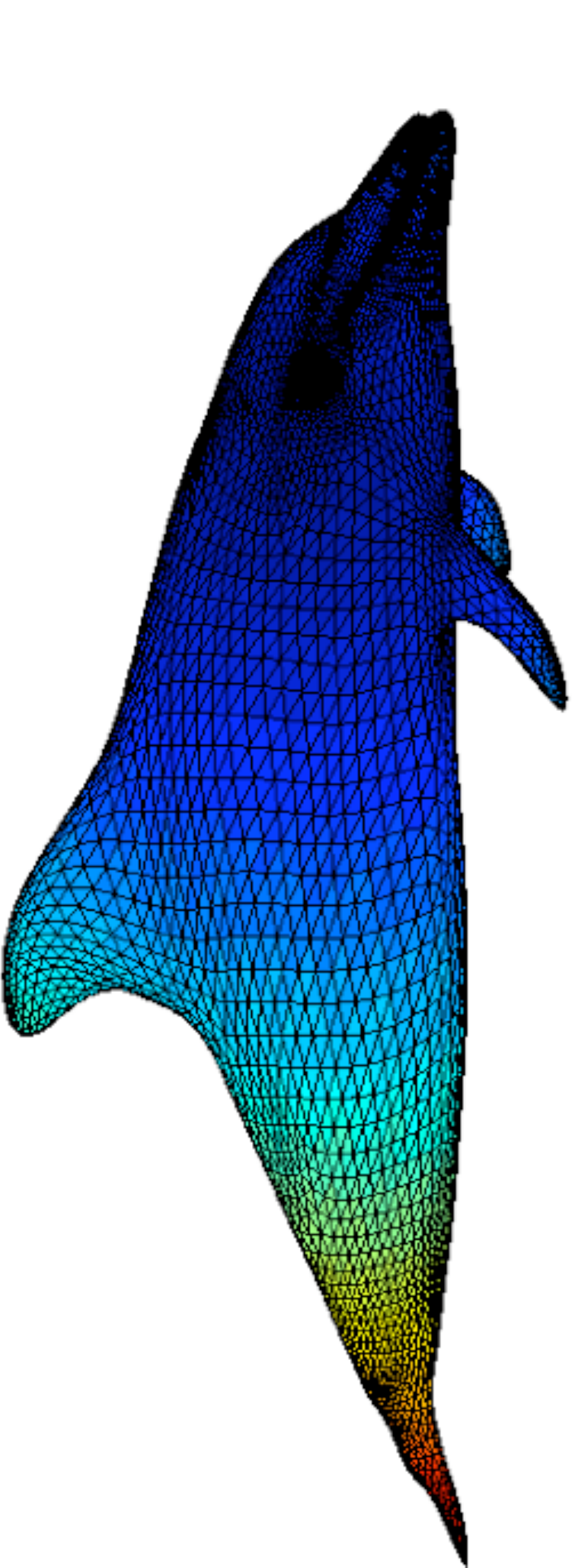}
   \hspace{0.5in}
\caption{3D models from The Princeton Shape Benchmark 
}\label{fig:Allobj}
\end{figure}

\begin{figure}[h]
\centerline{
\begin{tabular}{c}
\epsfig{figure=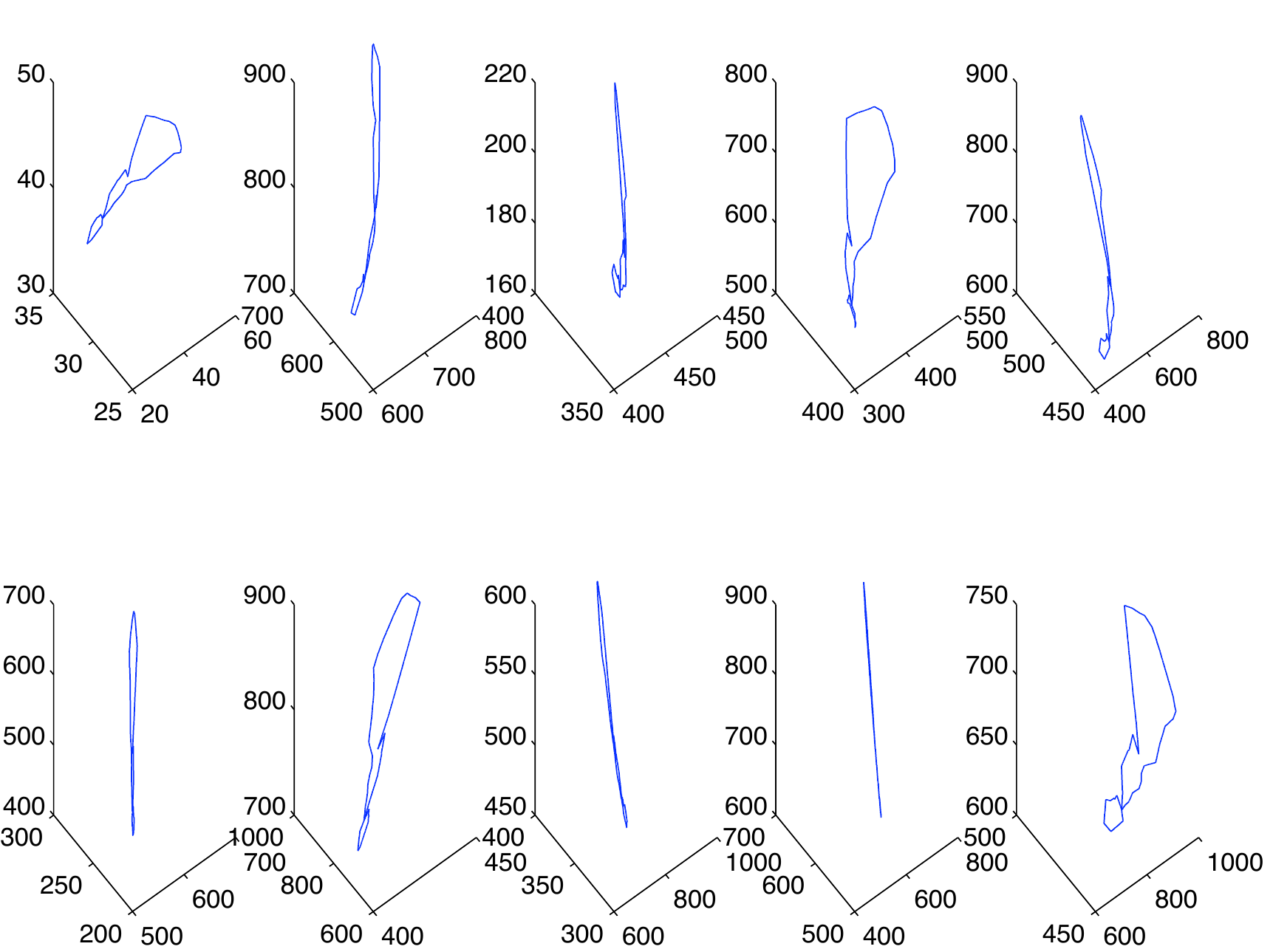,height=2in}
\end{tabular}}
\caption{{\it \small A curve and 9 variations of it under affine
transformation}} \label{fig:affinefigures}
\end{figure}

The Princeton Shape Benchmark \cite{princeton05} provides a
repository of 3D models. A subset of three models are shown in
Figure~\ref{fig:Allobj}. We extract a total of  100 characteristic
curves, and each of them are re-sampled to 5000 points with the same
arch-length. We applied to each curve 9 randomly generated 3D
special affine transformations as shown on
(Figure~\ref{fig:affinefigures}). To make this problem even more
challenging and to illustrate the noise sensitivity of the proposed
approach, gaussian noise with distribution $N(0,\sigma^{2})$ is
added to each of the variations. We therefore obtain a
classification set of 900 curves that has to be separated into 100
equivalent classes under affine transformations. The training set
consists of 100 original curves without any noise and
transformation. The discrimination power and sensitivity to noise
are analyzed using the error rate of classification.  We implemented
a Nearest Neighbor (NN) Classifier in a Euclidean Space using
Euclidean distance. In order to illustrate the advantages of the
signature, we design two experiments. The first experiment uses a
common parametrization for both the training and testing, while in
the second experiment, we choose two different parametrizations
(samplings) for the testing data.

\subsection{Experimental Results}

\begin{figure}
 \centerline{
\begin{tabular}{cc}
   \includegraphics[height=2 in]{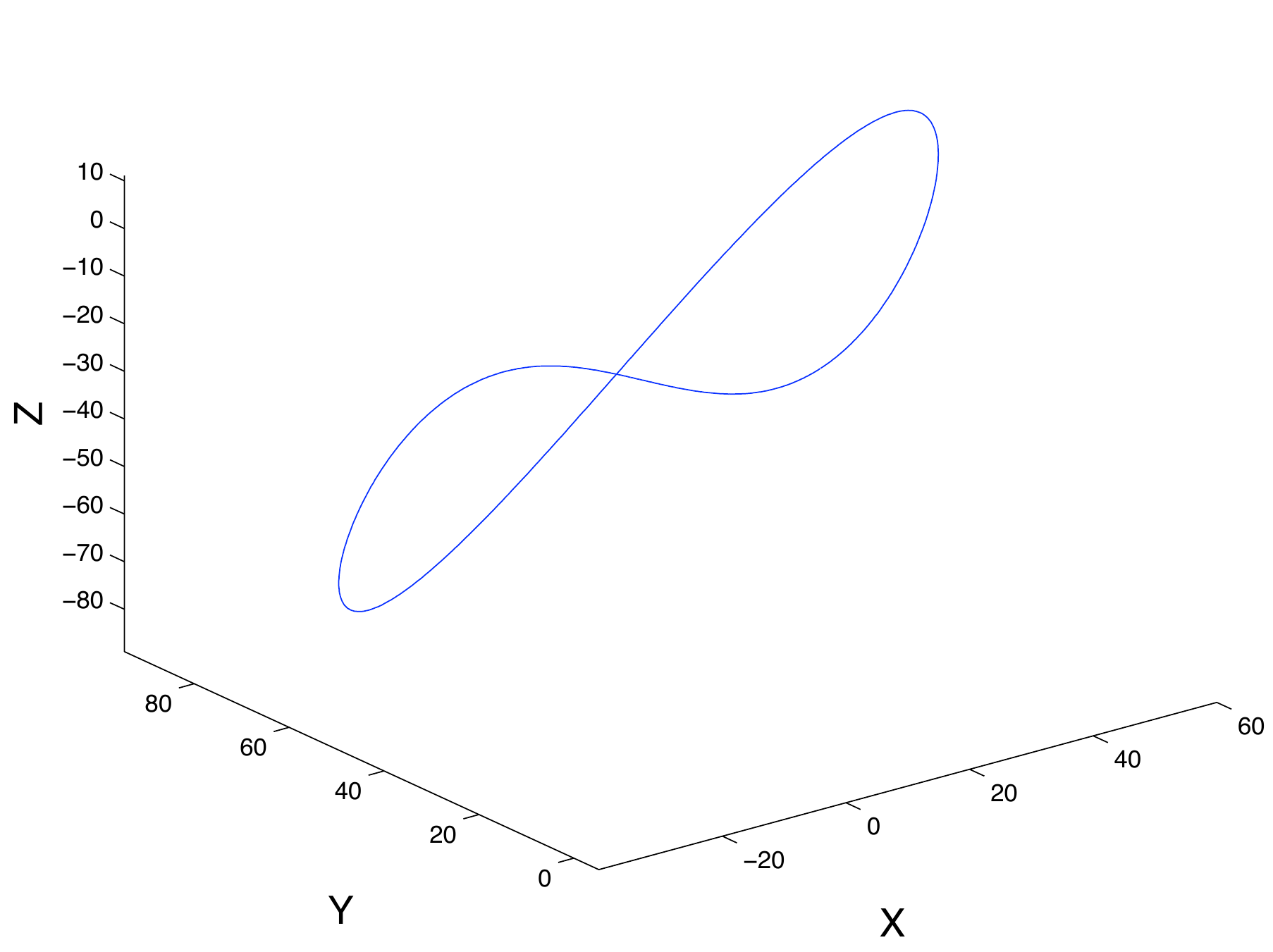} &
   \includegraphics[height=2 in]{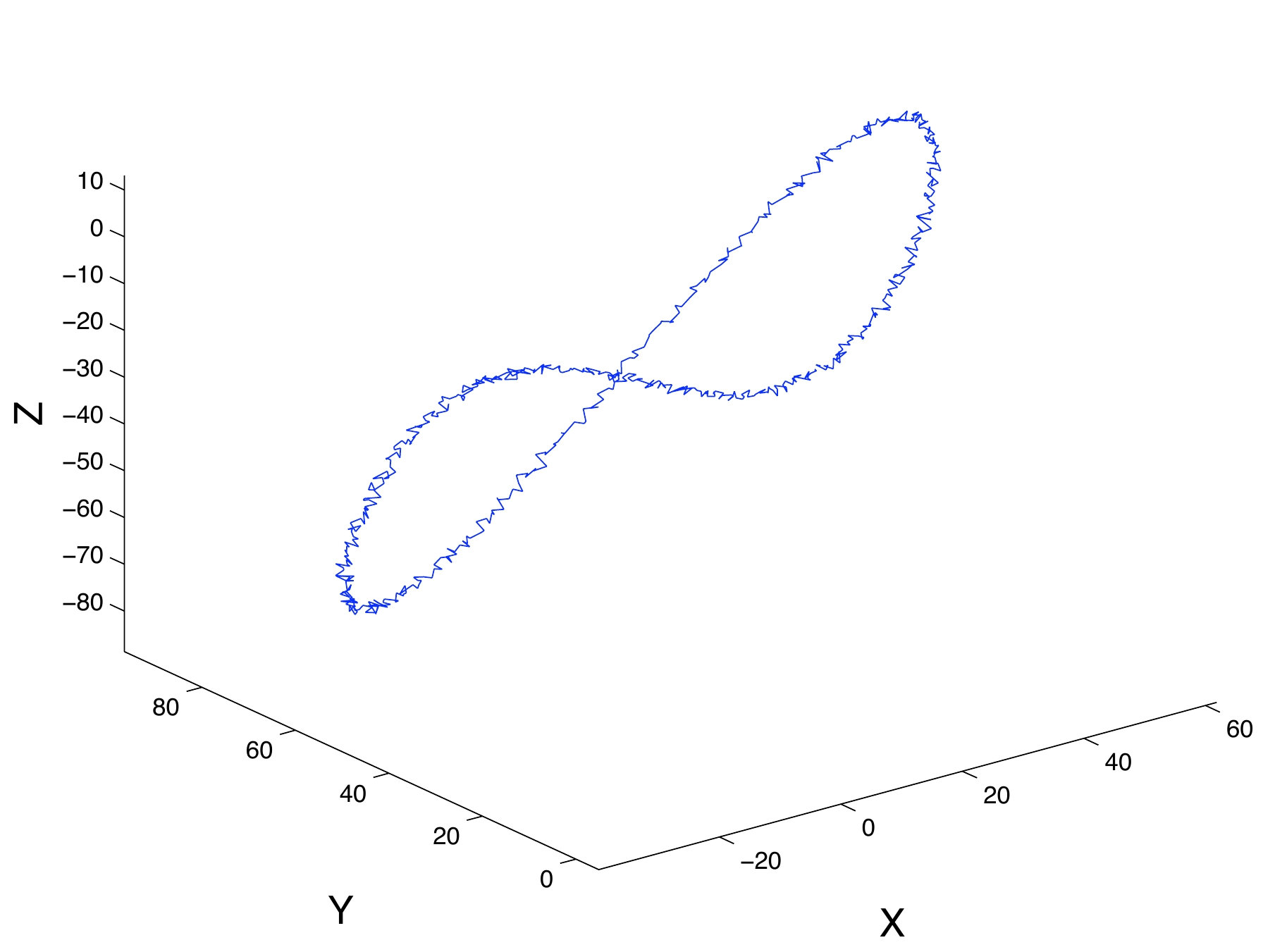}\\
    (a)&(b)\\
   \end{tabular}}
\caption{(a) A spatial curve without noise (b) with Gaussian Noise
$N(0,4)$, 
}\label{fig:curvewithnoise}
\end{figure}

Three experiments are carried out with different noise variance,
namely $\sigma=2$ (Fig. \ref{fig:curvewithnoise}), $\sigma=1$, and
$\sigma=0.5$. The error rates of the three different sigma settings
with the same parametrization are shown in Table.I.

\begin{table}
\center{
 \caption{Classification error rate with same parametrization, same initial points}
\begin{tabular}{|c|c|c|c|c|}
   \hline
  Noise variance    & $J_{1}$ & $J_{2}$ & Global signature&Local signature\\
  \hline
   $\sigma=0.5$  & \concept{0.0022} & 0.0472 & 0.06&0.07
 \\
   $\sigma=1$    & \concept{ 0.04}  & 0.12& 0.15 & 0.17\\
   $\sigma=2$    &  \concept{0.0789} & 0.2233& 0.28&0.32\\
   \hline
 \end{tabular}
} \label{table:errorrate1}
\end{table}

\begin{table}
\center{
 \caption{Classification error rate with different parametrization, same initial points}
\begin{tabular}{|c|c|c|c|c|}
\hline
   Noise variance    & $J_{1}$ & $J_{2}$ & Global signature&Local signature\\
 \hline
   $\sigma=0.5$  &\black{ 0.42 }& \black{0.61} & \concept{0.06} &\concept{0.07}
 \\
   $\sigma=1$    & \black{ 0.48}  &\black{ 0.70}&\concept {0.15}&\concept{0.17}\\
   $\sigma=2$    &  \black{0.56 }& \black{0.83}& \concept{0.28}&\concept{0.32}\\
   \hline
\end{tabular}
} \label{table:errorrate2}
\end{table}

\begin{table}
\center{
 \caption{Classification error rate with different parametrization, different initial points}
\begin{tabular}{|c|c|c|c|c|}
\hline
 Noise variance    & $J_{1}$ & $J_{2}$ & Global signature&Local signature\\
 \hline
   $\sigma=0.5$  &\black{ 0.87} & \black{0.95} & \black{0.95}&\concept{0.07}\\
   $\sigma=1$    &  \black{0.91}  & \black{0.97}& \black{0.97}&\concept{0.17}\\
   $\sigma=2$    &  \black{0.94} & \black{0.98}& \black{0.98}&\concept{0.32}\\
  \hline
\end{tabular}
} \label{table:errorrate3}
\end{table}

In Table I, both the integral invariants and the signatures perform
well as indicated by the error rates. For comparison, the classical
differential invariants have a classification error rate more then
80$\%$, which makes the differential invariants practically useless.
Since the order of integral variables involved in $J_{2}$ is higher
than $J_{1}$, as well as the explicit form of $J_{2}$ is more
complicated than $J_{1}$, the performance of $J_{2}$ is not as good
as that of $J_{1}$. The global signature is constructed with both
$J_{1}$ and $J_{2}$, and the local signature is based on $J_2$ and
$J_3$. The performance of these signatures is therefore slightly
worse than $J_{2}$.

If the parameterizations are not the same, the plots of invariants
$J_{1}$ and $J_{2}$ with respect to a parameter can not be used for
a classification purpose as illustrated in table II. Even with the
lowest noise variance, the error rates are more than 0.4 for $J_1$
and 0.6 for $J_2$. However, neither the global signature nor local
signature are affected.

If we make an arbitrary selection of the initial points, both
individual invariants ($J_1$ and $J_2$) and global signature have
poor performance as shown in Table III. Only local signature may be
used to characterize a curve.

As a conclusion, if the training data and testing data have similar
parameterization and same initial point, either invariants or
signatures may be used. Under different parametrization, the global
signature is the best choice. With an unknown starting point, the
local signature is the only solution.

\section{CONCLUSIONS}
In this paper, we presented explicit  formulas for affine integral
invariants for plane and space curves in terms of Euclidean
invariants. Based on the invariants, we constructed  signatures for
special affine groups and full affine groups for these curves.
Although we have focused here on a more complex case of affine
transformations, the integral Euclidean invariants, presented here,
can be used to classify curves under Euclidean
transformations \cite{feng07ssp}.

 { Integral invariants are functions of the parameter, and hence depend on the parameterization}.
 Global integral signatures provide a {classification method }
{independent of parameterization (curve sampling)}.  {A global
integral signature depends on the choice of the initial point. }
Local integral signatures provide a {classification method
independent of the choice of the initial point, they can hence be
used on images with occlusions  and for comparing fragments of
contours}. They are slightly more sensitive to noise than global
signatures.

As an experiment, a classification of characteristic
curves of 3D objects, subjected to random affine transformation
and noise, was conducted by using individual  invariants and global and local signatures.
Integral invariants allow us to perform {noise tolerant classification method for curves with respect to the affine transformations} (for comparison: differential invariants give 80\% error rate).

\section{APPENDIX: derivation of invariants}

\subsection{Cross-section and moving frame map}
Building on the works \scite{Gr74, jensen77,Green78}, Fels and
Olver \cite{FO99} generalized Cartan's normalization procedure
\cite{C35}, and proposed  a  general algorithm for computing
invariants.  The Fels-Olver algorithm relies on a map $\rho\colon
S\to G$   with an \emph {equivariant} property:
\eq{equiv}{\rho(g\cdot s)=\rho(s)\cdot g^{-1},\,\forall g\in G,
\forall s\in S.} From Theorem~4.4 in \scite{FO99}, it follows that
such map exists if and only if the action of $G$ is free  and, in
addition, there  exists a global \emph {cross-section}, i.e a
subset $\sva \subset S$ that intersects each orbit $O_s$ at a
unique point. Indeed, under the above assumption the map $\rho$
may  be defined by the condition $\rho(s)\cdot s\in \sva$. Then
$\rho(s)\cdot s=\rho(g\cdot s)\cdot (g\cdot s) $ is the unique
point of the intersection of $O_s$ and $\sva$. From the freeness
it follows  that  $s$ may be ``cancelled" and hence the condition
\Eq{equiv} is satisfied.

If $G$ is a  Lie group acting smoothly on $\R^n$ and  both
$S\subset\R^n$ and $\sva\subset S$  are smooth submanifolds, then
$\RR^n$-coordinate components of the
 projection $\iota(s)=\rho(s)\cdot s\colon S\to\sva$ are smooth invariant
 functions, called normalized invariants. Normalized invariants contain a
  maximal set of functionally independent invariants, and have a
  \emph{replacement property}, which allows us to rewrite any invariant in
  terms of them by simple substitution \cite{FO99, hk:jsc, hk:focm}.

Although, a global smooth cross-section does not always exist, a
local smooth cross-section\footnote{A local cross-section is
defined on an open subset of $U\subset S$ and $\forall s\in U$
intersects  each connected component  of $O_s\cap U$ at a unique
point.} passing through every point of $S$  may be found for every
semi-regular action.\footnote{ An action of $G$ is called \emph{semi-regular} if all orbits have the same dimension.} The  freeness assumption can be
also relaxed to a semi-regularity assumption. With these weaker
assumptions the above method can be used to construct local
invariants \cite{FO99,hk:focm}\footnote{A function $f$, defined on an open subset $U$
of $S$, is a \emph{local invariant} if $\forall s\in U$ there
exists an open neighborhood $G_s$ of $e\in G$ s.t.~condition
\Eq{inf} is  satisfied for all $g\in G_s$. }

For  algebraic groups acting  on algebraic varieties,  a purely
algebraic counterpart of the Fels-Olver construction was obtained in
\scite{hk:jsc, hk:focm}. The algebraic method can be combined with
the  inductive approach described below. In some particular
examples,  including the 3D example presented here, the computation
based on  the moving frame map $\rho$ turns out to be more
practical.

When the group $G$ is of relatively large dimension, computation of
invariants by either a geometric or algebraic approach becomes
challenging. In \scite{kogan03} two modifications of the moving
frame method were proposed to simplify the computation by splitting
it into two steps: invariants of a subgroup $A\subset G$ are first
computed, and then invariants of the entire group are constructed
 in terms of those. For the problem at hand, we use one of these
modifications, called the inductive approach, which is applicable
when a group factors into a product
 of two subgroups.

\subsection{ Inductive approach}
\begin{df} A group $G$ \emph{ factors as a product} of its subgroups $A$ and
$B$ if for any $g\in G$ there are $a\in A$ and $b\in B$ such that $g=ab$.
\end{df}
We write  $G=A\cdot B$. If in addition $A\cap B=e$, then for each
$g\in G$ there are
{\it unique} elements $a\in A$ and $b\in B$ such that $g=ab$.

Theorem $G=B\cdot A$, such that  $A\cap B$ is discrete,  and $G$ acts freely on a manifold  $S$ then $\forall s\in S$ there exists a local cross-section $\CK_A$, containing $s$, invariant under the action of the subgroup $B$. From Lemma~4.7 in \scite{kogan03} it follows that invariants of $G$ can be constructed from the invariants of $A$ using the following method.

{\bf Inductive method:}

\begin{enumerate}
 \item \vskip-2mm Restrict the $G$-action to $A$. Find a local cross-section $\sva_A\subset S$ for the action of $A$ which is invariant under the action
of $B$.
\item  Construct a moving frame map $\rho_A\colon \zva\to A$ defined by the condition    $\rho_A(s)\cdot s \in \sva_A,\,\forall s\in S$, by solving the corresponding equations. Composition of coordinate functions  with the projection  $\iota(s)=\rho_A(s)\cdot s\colon S\to\sva_A$ are invariant with respect to the action of $A$.

\item Restrict the action of $G$ to the action of its subgroup $B$ on the invariant subset $\sva_A$ and choose  a local cross-section $\sva_B\subset \sva_A$.
\item  Construct a moving frame map $\rho_B\colon \sva_A \to B$ defined by the condition    $\rho_B(s)\cdot z \in \sva_B, \,\forall z\in \sva_A$, by solving the corresponding equations. \comment{ We look at the ideal $<W-\alpha(b,w),\sva_A(w),\sva_B(W), B>\cap\K(w)[W]$. Substitute $w$ with replacement $A$-invariants obtained from the ideal $<Z-\alpha(a,z),\sva_A(Z),A>\cap \K(z)[Z]$$. Look at the simple example e.g Faugeras example, or even something more basic to see how it fits together. }
\item The $G$-moving frame map $\rho\colon S\to G$  is defined by $\rho(s)=\rho_B(\rho_A(s)\cdot s) \rho_A$, and $G$-invariants  are the coordinate components of   $\rho(s)\cdot s=\rho_B(\rho_A(s)\cdot s) \cdot(\rho_A (s)\cdot s)=\rho_B(\iota_A(s)) \cdot\iota_A (s)$.
\end{enumerate}

\subsection{ Integral affine invariants for curves in 2D}

We have a product decomposition  $SL(2)=B\cdot A$, where
$B=\left\{\left.\left(\begin {array}{cc}
b_{11} & b_{12}\\
0   &\frac 1 {b_{11}}
\end {array} \right)\right| b_{11}>0\right\}
$
 and   $A=SO(2)$ is a group of rotations.
The intersection  $B\cap A=\{e\}$, and therefore we  can apply the inductive
method as follows.
\begin{enumerate}
\item We  restrict the  $SL(2)$-action \Eq{2D} to  the subgroup  $\SO(2)$ of the rotation matrices
by setting
$a_{11}=\cos\phi,a_{12}= -\sin\phi, a_{21}=\sin\phi,
a_{22}=\cos\phi$.
A subset  $\sva_A$  defined by  conditions, $Y=0, X>0$ serves as a
cross-section   on an open subset of the integral jet bundle. Moreover $\sva_A$  is invariant under the restriction of \Eq{2D} to subgroup $B$.

\item The corresponding  moving frame map  $\rho_A(s)= \left(\begin {array}{cc}
\frac X {\sqrt{X^2+Y^2}}  & \frac Y {\sqrt{X^2+Y^2}}\\
-\frac Y {\sqrt{X^2+Y^2}} &\frac X {\sqrt{X^2+Y^2}}
\end {array} \right)
$
is obtained by solving the equation $\br Y=0$ with the condition $\br X>0$ (see \Eq{SO2inv}).
 The projection  $\iota_A\colon \R^5\to \sva_A$, obtained by substitution $\rho_A$ into \Eq{2D},  produces a point  whose coordinates are invariant under the action of $\SO(2)$. Non-constant normalized invariants are given by \Eq{SO2inv} and $Y_\SE=0$ is the  remaining constant invariant.
 \item We  now restrict the action \Eq{2D}  to the action of a subgroup  $B
 $ on an invariant subset $\sva_A$.  We obtain the following transformations.
 \eq{2DB}{\begin{array}{lcl}
\br {X_A}  = b_{11} X_A,& \quad&
 \br{Y^{[1,0]}_A}=Y^{[1,0]}_A,\\
 \br{Y^{[1,1]}_A}=\frac 1 {b_{11}}Y^{[1,1]}_A,&\quad&
  \br{Y^{[2,0]}_A}=b_{11}Y^{[2,0]}_A+2b_{12}Y^{[1,1]}_A\\
  \br{Y^{[1,2]}_A}=\frac 1 {b_{11}^2}Y^{[1,2]}_A,&\quad&\br{Y^{[2,1]}_A}=Y^{[2,1]}_A+2\frac {b_{12}}{b_{11}}Y^{[1,2]}_A,\\ 
  \br{Y^{[3,0]}_A}=b_{11}^2Y^{[3,0]}_A+3b_{12}^2Y^{[1,2]}_A+3 b_{11}b_{12}Y^{[2,1]}_A.\\
\end{array}}
A subset  $\sva_B\subset\sva_A$ defined by the
equations $ X_A=1,\quad Y^{[2,0]}_A=0$ serves as a cross-section on the subset of
$\sva_A$, where $Y^{[1,1]}_A\neq 0$.

\item This leads to the moving frame map $\rho_B(s)= \left(\begin {array}{cc}
 \frac 1 {X_A}  & -\frac{Y^{[2,0]}_A}{2\,Y^{[1,1]}_A X_A}\\
0 &{X_A}
\end {array} \right).
$
 The projection  $\iota_B\colon \sva_A\to \sva_B$, defined by  $\iota_B(s)=\rho_B(s)\cdot s$, produces a point  with coordinates
  \eq{inv2B}{\begin{array}{lcl}
 {X_B}  = 1,& \quad&
 {Y^{[1,0]}_B}=Y^{[1,0]}_A,\\
 {Y^{[1,1]}_B}=X_AY^{[1,1]}_A,&\quad&
  {Y^{[2,0]}_B}=0,\\
  {Y^{[1,2]}_B}=X_A^2Y^{[1,2]}_A,&\quad&{Y^{[2,1]}_B}=Y^{[2,1]}_A- \frac{Y^{[2,0]}_A}{Y^{[1,1]}_A }Y^{[1,2]}_A,\\
  {Y^{[3,1]}_B}=\frac 1{X_{A}^2}\left(Y^{[3,0]}_A+\frac 3 2 \frac{Y^{[2,0]}_A}{Y^{[1,1]}_A }Y^{[2,1]}_A+\frac 3 4\left(\frac{Y^{[2,0]}_A}{Y^{[1,1]}_A} \right)^2Y^{[1,2]}_A\right).&&
\end{array}}
 invariant under the $B$-action \Eq{2DB} on $\sva_A$.

\item Replacing coordinates $ Y^{[i,j]}_A$ with the corresponding  $ Y^{[i,j]}_\SE$ given by \Eq{SO2inv} produces   independent invariants \Eq{SL2inv}.
\end{enumerate}

\subsection {Affine integral invariants for curves in 3D}

We have a product decomposition  $SL(3)=B\cdot A$, where
$B=\left\{\left(\begin {array}{ccc}
b_{11} & b_{12}& b_{13}\\
0 & b_{22} & b_{23}\\
0 & 0  &\frac {1} {b_{11}b_{22}}
\end {array} \right)\left|b_{11}>0,b_{22}\neq 0\right.\right\}
$ and $A=SO(3)$ is a group of rotations. The intersection  $B\cap
A=\{e\}$ is trivial. We again follow the steps of the inductive
method.
\begin{enumerate}
\item We restrict the $\SL(3)$ to the action of $\SO(3)$ whose elements can be represented as the product of three rotations:
$$\left( \begin {array}{ccc}
1&0&0\\\noalign{\medskip}0&{\cos\psi}&-{\sin\psi}\\\noalign{\medskip}0&{\sin\psi}&{\cos\psi}\end
{array} \right)
 \left( \begin {array}{ccc} {\cos\phi}&0&{\sin\phi}\\\noalign{\medskip}0&
1&0\\\noalign{\medskip}{-\sin\phi}&0&{\cos\phi}\end {array}
\right) \left( \begin {array}{ccc} {\cos\theta}&{
-\sin\theta}&0\\\noalign{\medskip}{ \sin\theta}&{\cos\theta}&0\\\noalign{\medskip}0&0&1\end {array} \right).$$
 A
subset $\sva_A$,  defined by  conditions, $Y=0,\,Z=0,\,Z_{011}=0, X>0$
serves as a cross-section on the open subset of the integral jet bundle where
$X^2+Y^2+Z^2>0$.  The cross-section $\sva_A$  is invariant under the action of $B$. \item
The corresponding  moving frame map  $\rho_A$ is obtained by
solving the equation $\br Y=0, \br Z=0, \br {Z^{[0,1,1]}}=0$ with the
condition $\br X>0$.  Explicitly \eq{}{\begin{array}{lll} \cos\theta=\frac
{X}{\sqrt {X^2+Y^2}},& \cos\phi={\frac {\sqrt
{{X}^{2}+{Y}^{2}}}{\sqrt {{X}^{2}+{Y}^{2}+{Z}^{2}}}},
 &\cos \psi =\frac {Z^{[0,2,0]}_R}{\sqrt {\left(Z^{[0,2,0]}_R\right)^{2}+4\,
\left(Z^{[0,1,1]}_R\right)^2}},\\
\sin\theta=-\frac {Y}{\sqrt { {X}^{2}+{Y}^{2} } },& \sin\phi=\frac
{Z}{\sqrt {{X}^{2}+{Y}^{2}+{Z}^{2} }},& \sin \psi =-2\, \frac
{Z^{[0,1,1]}_R} {\sqrt {\left(Z^{[0,2,0]}_R\right)^{2}+4\,
\left(Z^{[0,1,1]}_R\right)^2}}.
\end{array}
}
 we ${Z^{[0,1,1]}_R}$ and $Z^{[0,2,0]}_R$ are given on the last page of the Appendix.
The corresponding set of $\SO(3)$ invariants is given by \Eq{SO3inv}.

\item We  now restrict the action $\SL(3)$-action  to the action of a subgroup  $B
 $ on an invariant subset of $\sva_A$.  We obtain the following transformations:
\begin{eqnarray*}
\br{X_A}&=& b_{11}\,{ X_A},\\
\br {Z^{[0,1,0]}_A}&=&\frac 1 {b_{11}}\,  Z^{[0,1,0]}_A,\\
\br{Z^{[1,0,0]}_A}& =&\frac 1 {b_{22}}\, Z^{[1,0,0]}_A+ \frac{b_{12}}{b_{11}b_{22}}\, Z_A^{[0,1,0]},\\
\br{Y_A^{[1,0,0]}}&=&b_{11}\, b_{22}\, Y^{[1,0,0]}- b_{13}\, b_{22}\, Z^{[0,1,0]}_A+ b_{11}
\, b_{23}\, Z^{[1,0,0]}_A+ b_{12}\, b_{23}\,
Z^{[0,1,0]}_A,\\
\br{Z^{[0,2,0]}_A}&=&\frac{b_{22}}{b_{11}}\, Z_A^{[0,2,0]},\\
\br{Z_A^{[1,0,1]}}&=& \frac{1}{b_{11}\, b_{22}^{2}} Z_A^{[1,0,1]}\\
\br{Z_A^{[1,1,0]}}&=&Z_A^{[1,1,0]}+\frac { b_{23}}{ b_{22}}\, Z_A^{[1,0,1]}+\frac{b_{12}}{ b_{11}}\, Z_A^{[0,2,0]},\\
\br{Y_A^{[1,0,1]}}&=&Y_A^{[1,0,1]}+\frac {b_{23}}{b_{22}}Z_A^{[1,0,1]}-\frac
{b_{12}}{2b_{11}}Z_A^{[0,2,0]}.
\end{eqnarray*}
A subset  $\sva_B\subset\sva_A$ defined by
equations $$
{Z_A^{[0,1,0]}}=1,\,{Z_A^{[1,0,0]}}=1,\,{Y_A^{[1,0,0]}}=1,\,{Z_A^{[0,2,0]}}=1,{Z_A^{[1,1,0]}}=1$$ is a cross-section on an open  subset of $\sva_A$.

\item The  corresponding moving frame map $\rho_B$ is
\begin{eqnarray*}
b_{11}&=&Z_A^{[0,1,0]},
\quad b_{12}=-\frac
{Z_A^{[0,2,0]}\,Z_A^{[1,0,0]}-Z_A^{[0,1,0]}}
{Z_A^{[0,2,0]}},\\
b_{{22}}&=&\frac
{Z_A^{[0,1,0]}}{Z_A^{[0,2,0]}},\quad b_{{23}}=\frac
{-Z_A^{[0,1,0]}Z_A^{[1,1,0]}+Z_A^{[0,2,0]}Z_A^{[1,0,0]}}{Z_A^{[0,2,0]}Z_A^{[1,0,1]}},\\
b_{{13}}&=&-\frac
{-{Z_A^{[0,1,0]}}^{2}Z_A^{[0,2,0]}Z_A^{[1,0,0]}+{Z_A^{[0,1,0]}}^{3}Z_A^{[
1,1,0]}-Z_A^{[0,2,0]}Z_A^{[1,0,1]}{Z_A^{[0,1,0]}}^{2}{
Y_A^{[1,0,0]}}+Z_A^{[1,0,1]}{Z_A^{[0,2,0]}}^{2}}
{Z_A^{[0,2,0]}Z_A^{[1,0,1]}{Z_A^{[0,1,0]}}^{2}}.
\end{eqnarray*}
The coordinate components of the projection  $\rho_B(s)\cdot s\colon \sva_A\to \sva_B$
\begin{eqnarray*}
 X_B&=& Z_A^{[0,1,0]}X_{A}\\
  Z_B^{[1,0,1]}& =&
\frac{Z_A^{[1,0,1]}{Z_A^{[0,2,0]}}^2}{{Z_A^{[0,1,0]}}^3},\\
Y_B^{[1,0,1]}&=&
\frac{2Y_A^{[1,0,1]}Z_A^{[0,1,0]}-2Z_A^{[0,1,0]}Z_A^{[1,1,0]}+3Z_A^{[0,2,0]}Z_A^{[1,0,0]}}{2Z_A^{[0,1,0]}}-\frac{1}{2}.
\end{eqnarray*}
are invariant under the action of $B$ on $\sva_A$.

\item Replacing coordinates $Y^{[i,j,k]}_A, Z^{[i,j,k]}_A$ with the corresponding  $Y^{[i,j,k]}_\SE, Z^{[i,j,k]}_\SE$ given by \Eq{SO3inv} produces   independent invariants \Eq{SL3inv}.

\end{enumerate}
The following auxiliary expressions were used in the paper where   $r=\sqrt{X^2+Y^2}, R=\sqrt{X^2+Y^2+Z^2}$:
\eqs{3DEuclidean1_1}{
\nonumber Z_{R}^{[0,1,0]}&=&-\frac{1}{2R}\left(XYZ-2\,X\,
Z^{[0,1,0]}+2\,Y\, Z^{[1,0,0]}-2\,Z\, Y^{[1,0,0]}\right)\\
\nonumber Z_R^{[1,0,0]}&=& -\frac{1}{2r}\left({X}^{2}Z-2\,X{
Z^{[1,0,0]}}+{Y}^{2}Z-2\,Y{ Z^{[0,1,0]}}\right)\\
\nonumber Y_{R}^{[1,0,0]}&=&-\frac{1}{2\,r\,R}\left(
Y{X}^{3}+X{Y}^{3}-2\,{Y}^{2}{ Y^{[1,0,0]}}-2\,ZY{ Z^{[1,0,0]}}-2
\,{X}^{2}{ Y^{[1,0,0]}}+2\,ZX{ Z^{[0,1,0]}}\right)\\
\nonumber Z_R^{[0,1,1]}&=&-\frac 1{6\,r\,R}
\left(2\,Y{X}^{3}{Z}^{2}-6\,{X}^{3}{
Z^{[0,1,1]}}+6\,{X}^{2}Y{ Z^{[1,0,1]}}-6\,{X}^{2}Z{
Y^{[1,0,1]}}\right. \\\nonumber& &\left.
-6\,X{
Z^{[0,1,1]}}\,{Y}^{2}+3\,XYZ{ Z^{[0,2,0]}}
+4 \,{Z}^{2}{Y}^{3}X+6\,XYZ{
X^{[1,0,1]}}-6\,X{Z}^{2}{ X^{[1,1,0]}}-6\,{Y}^{2}Z{
Z^{[1,1,0]}}
\right. \\\nonumber& &\left.
-6\,{Y}^{2}Z{ Y^{[1,0,1]}}+6\,{Y}^{3}{
Z^{[1,0,1]}}-3\,{Z}^{2}Y{ X^{[0,2,0]}}\right)\\
\nonumber Z_R^{[0,2,0]}&=& -\frac{1}{3\,r\,R}\left(-3\,{X}^{2}{
Z^{[0,2,0]}}-2\,{Y}^{2}{X}^{2}Z+6\,XY{ Z^{[1,1,0]}} +3\,XZ{
X^{[0,2,0]}}+6\,{Y}^{2}{ X^{[1,0,1]}}-6\,ZY{ X^{[1,1,0]}}\right)\\
\nonumber Z_R^{[1,0,1]}&=& \frac 1{6\,r^2\,R}\left(-4\,{Z}^{2}{X}^{4}+6\,{
Z^{[1,0,1]}}\,{X}^{3}+6\,{X}^{2}Y{ Z^{[0,1,1]}
}-2\,{X}^{2}{Z}^{2}{Y}^{2}+6\,Z{X}^{2}{ X^{[1,0,1]}}-6\,XYZ{
Z^{[1,1,0]}}
\right. \\\nonumber& &\left. +6\,X { Z^{[1,0,1]}}\,{Y}^{2}-3\,{Y}^{2}Z{
Z^{[0,2,0]}}+6\,{Y}^{3}{ Z^{[0,1,1]}}-{Z}^{2} {Y}^{4}\right)\\
\nonumber Z_R^{[1,1,0]}&=&\frac 1{6\,r^2\,R^2}
\left(6\,{Z}^{2}XY{
X^{[1,0,1]}}-6\,ZX{Y}^{2}{ X^{[1,1,0]}}-3\,Z{X}^{2}Y{
X^{[0,2,0]}}+6\,{Z}^{2}{Y}^{2}{
Y^{[1,0,1]}}
\right. \\\nonumber& &\left.
-4\,Z{X}^{5}Y+Z{Y}^{5}X+6\,{X}^{4 }{ Z^{[1,1,0]}}
-6\,{Y}^{4}{ Z^{[1,1,0]}}-6\,ZY{
Z^{[1,0,1]}}\,{X}^{2}+6\,ZX{ Z^{[0,1,1]}}\,{Y}^{2}
\right. \\\nonumber& &\left.+3\,{Z}^{2}XY{
Z^{[0,2,0]}}-6\,Z{X}^{3}{ X^{[1,1,0]}}-3\,{X}^{3} {Y}^{3}Z
 +12\,{X}^{3}Y{ X^{[1,0,1]}}+12\,X{Y}^{3}{
X^{[1,0,1]}}
\right. \\\nonumber& &\left.
-6\,Z{Y}^{3}{  Z^{[1,0,1]}}
+6\,{X}^{3}Y{
Z^{[0,2,0]}}+6\,X{Y}^{3}{ Z^{[0,2,0]}}+6\,Z{X}^{3}{ Z^{[0,1,1]}}\right. \\\nonumber& &\left.
-6\,{X}^{3}{Z}^{3}Y+6\,{X}^{2}{Z}^{2}{
Z^{[1,1,0]}}+6\,{X}^{2}{Z}^{2} { Y^{[1,0,1]}}-3\,Z{Y}^{3}{
X^{[0,2,0]}}-3\,{Z}^{3}{Y}^{3}X\right)%
\\
\nonumber Y_R^{[1,0,1]}&=&- \frac 1{6\,r^2\,R^2} \left(-6\,{X}^{4}{
Y^{[1,0,1]}}-12\,ZX{Y}^{2}{ X^{[1,1,0]}}+6\,{Z}^{2} {Y}^{2}{
Y^{[1,0,1]}}-6\,{Y}^{4}{ Z^{[1,1,0]}}
-6\,Z{X}^{2}Y{X^{[0,2,0]}}
\right. \\\nonumber& &\left.
-6\,{Y}^{ 2}{ Z^{[1,1,0]}}\,{X}^{2}
-12\,ZY{
Z^{[1,0,1]}}\,{X}^{2}+6\,{Y}^{2}{Z}^{2}{
Z^{[1,1,0]}}-12\,{Y}^{2}{X}^{2}{ Y^{[1,0,1]}}-6\,{Y}^{4}{
Y^{[1,0,1]}}
\right. \\\nonumber& &\left.
-6\,Z{Y}^{3}{  X^{[0,2,0]}}-12\,Z{X}^{3}{ X^{[1,1,0]}}
-3\,{Z}^{3}{Y}^{3}X+9\,{X}^{3}{Y}^{3}Z
+4\,Z{X}^{5}Y-12\,Z{Y}^{3}{ Z^{[1,0,1]}}
\right. \\\nonumber& &\left.
+12\,Z{X}^{3}{
Z^{[0,1,1]}}+3\,X{Y}^{3 }{ Z^{[0,2,0]}}+5\,Z{Y}^{5}X-3\,{Z}^{2}XY{ Z^{[0,2,0]}}+12\,ZX{
Z^{[0,1,1]}}\,{Y}^ {2}-6\,{Z}^{2}XY{ X^{[1,0,1]}}
\right. \\\nonumber& &\left.+6\,X{Y}^{3}{
X^{[1,0,1]}}+6\,{X}^{3}Y{ X^{[1,0,1]}} +6\,{X}^{2}{Z}^{2}{
Y^{[1,0,1]}}+3\,{X}^{3}Y{ Z^{[0,2,0]}}\right)
\\%
\nonumber X_R^{[1,1,0]}&=&\frac 1{6\,r\,R^2}\left(-{Y}^{5}X+3\,{Z}^{2}{Y}^{3}X-6\,Z{Y}^{2}{ Y^{[1,0,1]}}-6\,Z{X}^{2}
{ Y^{[1,0,1]}}-6\,Z{Y}^{2}{
Z^{[1,1,0]}}-2\,{X}^{5}Y
\right. \\\nonumber& &\left.
+6\,{X}^{3}{ X^{[1,1,0]}}+3\,X YZ{ Z^{[0,2,0]}}+3\,{X}^{2}Y{
X^{[0,2,0]}}+6\,XYZ{ X^{[1,0,1]}}+6\,X{Y}^{2}{ X^{[1,1,0]}}\right. \\\nonumber& &\left.
+3\,{Y}^{3}{
X^{[0,2,0]}}-3\,{X}^{3}{Y}^{3}-6\,{Z}^{2}Y{ Z^{[1,0,1]}}+6\,
{Z}^{2}X{ Z^{[0,1,1]}}\right),\\
 \nonumber X_R^{[1,0,1]}&=&\frac 1{6\,r\,R}
\left(-2\,Z{X}^{4}+6\,{X}^{2}{ X^{[1,0,1]}}+2\,{Y}^{2}{X}^{2}Z+2\,{
X}^{2}{Z}^{3}-6\,XY{ Z^{[1,1,0]}}-6\,ZX{ Z^{[1,0,1]}}
\right. \\\nonumber& &\left.
-6\,ZY{
Z^{[0,1,1]}}+2\,{Y}^{ 2}{Z}^{3}+{Y}^{4}Z-3\,{
Z^{[0,2,0]}}\,{Y}^{2}\right),\\
\nonumber X_R^{[0,2,0]}&=& -\frac 1{3\,r^2\,R}
\left(-{X}^{4}{Y}^{2}-{Y}^{4}{X}^{2}-3\,X{Y}^{2}{ X^{[0,2,0]}}-3\,
{X}^{2}{Z}^{2}{Y}^{2}+6\,{Y}^{2}Z{ X^{[1,0,1]}}+6\,XYZ{ Z^{[1,1,0]}}
\right. \\\nonumber& &\left.-3\,Z{X}^{ 2}{
Z^{[0,2,0]}}+6\,{X}^{2}Y{ X^{[1,1,0]}}-3\,{X}^{3}{
X^{[0,2,0]}}+6\,{Y}^{3}{  X^{[1,1,0]}}\right).
}

\newpage
\bibliographystyle{plain}
\newcommand{\SortNoop}[1]{}

\end{document}